\documentclass[11pt]{article}
\usepackage{fullpage}

\usepackage{times}
\usepackage{epsfig}
\usepackage{graphicx}
\usepackage{amsmath} 
\usepackage{amssymb}
\usepackage{diagbox}
\usepackage{pifont}

\usepackage{threeparttable}

\usepackage[pagebackref=true,breaklinks=true,colorlinks,bookmarks=false]{hyperref}

\usepackage{comment}
\usepackage{booktabs}       
\usepackage{multirow}

\usepackage{subfigure}
\usepackage{soul}

\usepackage{enumitem}

\usepackage{bbding}
\usepackage{placeins}

\usepackage{amsmath}
\usepackage{amsfonts}
\usepackage{amssymb}

\usepackage{wrapfig}

\newcommand{\RN}[1]{%
	\textup{\lowercase\expandafter{\it \romannumeral#1}}%
}

\usepackage{xcolor}
\definecolor{mygreen}{HTML}{3cb44b}
\definecolor{skyblue}{HTML}{beffff}
\definecolor{lightgreen}{HTML}{90ee90}

\newcommand{\ea}[0]{\emph{et al. }}

\newcommand{\beq}{\vspace{0mm}\begin{equation}}
\newcommand{\eeq}{\vspace{0mm}\end{equation}}
\newcommand{\beqs}{\vspace{0mm}\begin{eqnarray}}
\newcommand{\eeqs}{\vspace{0mm}\end{eqnarray}}
\newcommand{\barr}{\begin{array}}
\newcommand{\earr}{\end{array}}

\newcommand{\Wmat}[0]{{{\bf W}}}

\newcommand{\hv}[0]{{\boldsymbol{h}}}

\newcommand{\qv}[0]{{\boldsymbol{q}}}

\newcommand{\vv}{\boldsymbol{v}}
\newcommand{\wv}{\boldsymbol{w}}
\newcommand{\xv}{\boldsymbol{x}}

\newcommand{\thetav}{\boldsymbol{\theta}}

\newcommand{\E}{\mathbb{E}}

\usepackage{caption}

\newcommand{\Lcal}{\mathcal{L}}

\newcommand{\Dcal}{\mathcal{D}}

\definecolor{Gray}{gray}{0.93}
\definecolor{Graylight}{gray}{0.95}
\definecolor{Grayheavy}{gray}{0.90}

\usepackage{colortbl}
\definecolor{Gray}{gray}{0.93}

\newcommand{\ie}[0]{\emph{i.e., }}
\newcommand{\etal}[0]{\emph{et al. }}

\newcommand{\oscar}{\textsc{Oscar}}
\newcommand{\short}{\textsc{Oscar+}}
\newcommand{\shortb}{\textsc{Oscar+}$_{\text{B}}$}
\newcommand{\shortl}{\textsc{Oscar+}$_{\text{L}}$}
\newcommand{\vinvl}{\textsc{VinVL}}
\newcommand{\vinvlb}{\textsc{VinVL}$_{\text{B}}$}
\newcommand{\vinvll}{\textsc{VinVL}$_{\text{L}}$}

\definecolor{MyPurple}{RGB}{111,0,255}

\usepackage{lipsum}

\newcommand\blfootnote[1]{%
  \begingroup
  \renewcommand\thefootnote{}\footnote{#1}%
  \addtocounter{footnote}{-1}%
  \endgroup
}

\begin{document}

\title{VinVL: Revisiting Visual Representations \\ in Vision-Language Models}

\author{
Pengchuan Zhang\textsuperscript{$\heartsuit\dagger$} 
\and Xiujun Li\textsuperscript{$\heartsuit\spadesuit\dagger$}
\and Xiaowei Hu\textsuperscript{$\heartsuit$} 
\and Jianwei Yang\textsuperscript{$\heartsuit$}
\and Lei Zhang\textsuperscript{$\heartsuit$} 
\and Lijuan Wang\textsuperscript{$\heartsuit$} 
\and Yejin Choi\textsuperscript{$\spadesuit$} 
\and Jianfeng Gao\textsuperscript{$\heartsuit$}
}

\maketitle

\blfootnote{\textsuperscript{$\heartsuit$}Microsoft Corporation
\hspace{10mm}
 \textsuperscript{$\spadesuit$}University of Washington
 \hspace{10mm}
 $\dagger$ indicates equal contributions.}

\begin{abstract}

This paper presents a detailed study of improving visual representations for vision language (VL) tasks and develops an improved object detection model to provide object-centric representations of images.
Compared to the most widely used \emph{bottom-up and top-down} model \cite{anderson2018bottom}, the new model is bigger, better-designed for VL tasks, and pre-trained on much larger training corpora that combine multiple public annotated object detection datasets. Therefore, it can generate representations of a richer collection of visual objects and concepts. While previous VL research focuses mainly on improving the vision-language fusion model and leaves the object detection model improvement untouched, we 
show that visual features matter significantly in VL models.
In our experiments we feed the visual features generated by the new object detection model into a Transformer-based VL fusion model \oscar~\cite{li2020oscar}, and utilize an improved approach \short\  to pre-train the VL model and fine-tune it on a wide range of downstream VL tasks. 
Our results show that the new visual features significantly improve the performance 
across all VL tasks, creating new state-of-the-art results on seven public benchmarks. Code, models and pre-extracted features are released at \url{https://github.com/pzzhang/VinVL}.

\end{abstract}


\section{Introduction}

Vision language pre-training (VLP) has proved effective for a wide range of vision-language (VL) tasks
~\cite{lu2019vilbert,tan2019lxmert,chen2019uniter,su2019vl,li2019visualbert,li2019unicoder,zhou2019unified,li2020oscar}. VLP typically consists of two stages: 
(1) an object detection model is pre-trained to encode an image and the visual objects in the image to feature vectors, and
(2) a cross-modal fusion model is pre-trained to blend text and visual features.
While existing VLP research focuses mainly on improving the cross-modal fusion model, this paper focuses on improving the object-centric visual representations and presents a comprehensive empirical study to demonstrate that visual features matter in VL models.

\begin{table*}[!ht]
\begin{center}
\resizebox{\linewidth}{!}{
\scriptsize
\begin{tabular}{c@{\hspace{3pt}}|c@{\hspace{5pt}}c@{\hspace{3pt}}|c@{\hspace{5pt}}c|c@{\hspace{5pt}}c@{\hspace{5pt}}c@{\hspace{6pt}}c@{\hspace{5pt}}|c@{\hspace{6pt}}c@{\hspace{6pt}}|c@{\hspace{6pt}}c@{\hspace{6pt}}c@{\hspace{6pt}}|c@{\hspace{6pt}}c@{\hspace{6pt}}c@{\hspace{3pt}}|c@{\hspace{6pt}}c}
\toprule
\multirow{2}{*}{Visual feature} & \multicolumn{2}{c|}{VQA} & \multicolumn{2}{c|}{GQA} & \multicolumn{4}{c|}{Image Captioning} & \multicolumn{2}{c|}{NoCaps} & \multicolumn{3}{c|}{Image Retrieval} & \multicolumn{3}{c|}{Text Retrieval} & \multicolumn{2}{c}{NLVR2} \\ 
& test-dev & test-std & test-dev & test-std & B@4 & M & C & S & C & S & R@1 & R@5 & R@10 & R@1 & R@5 & R@10 & dev & test-P \\ \midrule
Anderson \ea \cite{anderson2018bottom} & $73.16$ & $73.44$ & $61.58$ & $61.62$ & $40.5$ & $29.7$ & $137.6$ & $22.8$ & $86.58$ & $12.38$ & $54.0$ & $80.8$ & $88.5$ & $70.0$ & $91.1$ & $95.5$ & $78.07$ & $78.36$ \\
\rowcolor{Graylight}  
Ours & $\bf 75.95$ & $\bf 76.12$ & $\bf 65.05$ & $\bf 64.65$ & $\bf 40.9$ & $\bf 30.9$ & $\bf 140.6$ & $\bf 25.1$ & $\bf 92.46$ & $\bf 13.07$ & $\bf 58.1$ & $\bf 83.2$ & $\bf 90.1$ & $\bf 74.6$ & $\bf 92.6$ & $\bf 96.3$ & $\bf 82.05$ & $\bf 83.08$ \\ 
\hline
$\Delta$ & $\bf 2.79\uparrow$ & $\bf 2.68\uparrow$ & $\bf 3.47\uparrow$ & $\bf 3.03\uparrow$ & $\bf 0.4\uparrow$ & $\bf 1.2\uparrow$ & $\bf 3.0\uparrow$ & $\bf 2.3\uparrow$ & $\bf 5.9\uparrow$ & $\bf 0.7\uparrow$ & $\bf 4.1\uparrow$ & $\bf 2.4\uparrow$ & $\bf 1.6\uparrow$ & $\bf 4.6\uparrow$ & $\bf 1.5\uparrow$ & $\bf 0.8\uparrow$ & $\bf 3.98\uparrow$ & $\bf 4.71\uparrow$ \\ 
\bottomrule
\end{tabular}
}
\end{center}
\vspace{-5mm}
\caption{
Uniform improvements on seven VL tasks by replacing visual features from Anderson \ea \cite{anderson2018bottom} with ours.
The NoCaps baseline is from VIVO~\cite{hu2020vivo}, and our results are obtained by directly replacing the visual features. The baselines for rest tasks are from \textsc{Oscar}~\cite{li2020oscar}, and our results are obtained by replacing the visual features and performing $\short$ pre-training. All models are BERT-Base size. As analyzed in Section~\ref{subsec:oscar_ablation}, the new visual features contributes 95\% of the improvement.}
\label{tab:peter_vinvl_comp}
\end{table*}

\begin{figure*}[t!]
\includegraphics[width=0.5\textwidth]{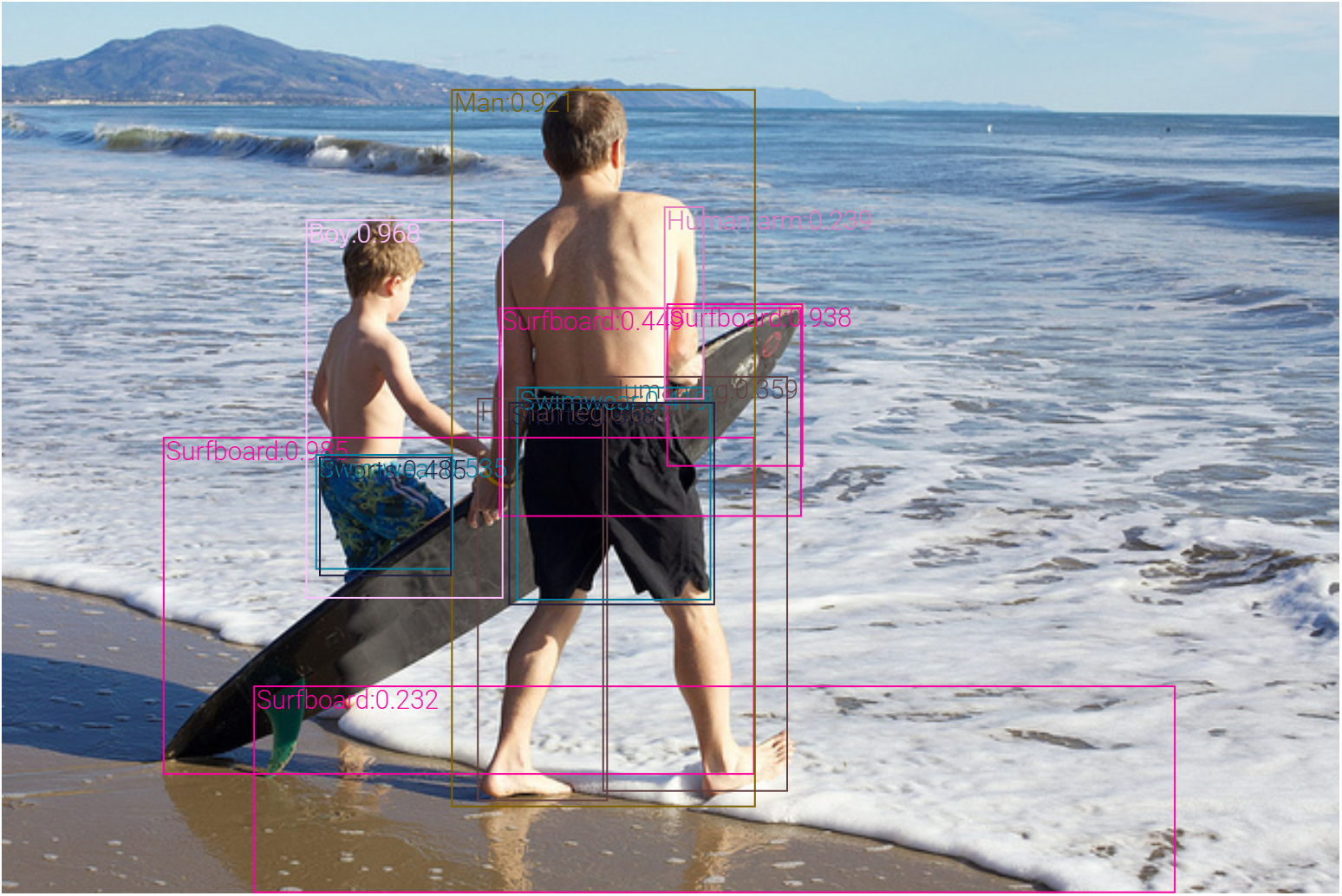}
\includegraphics[width=0.5\textwidth]{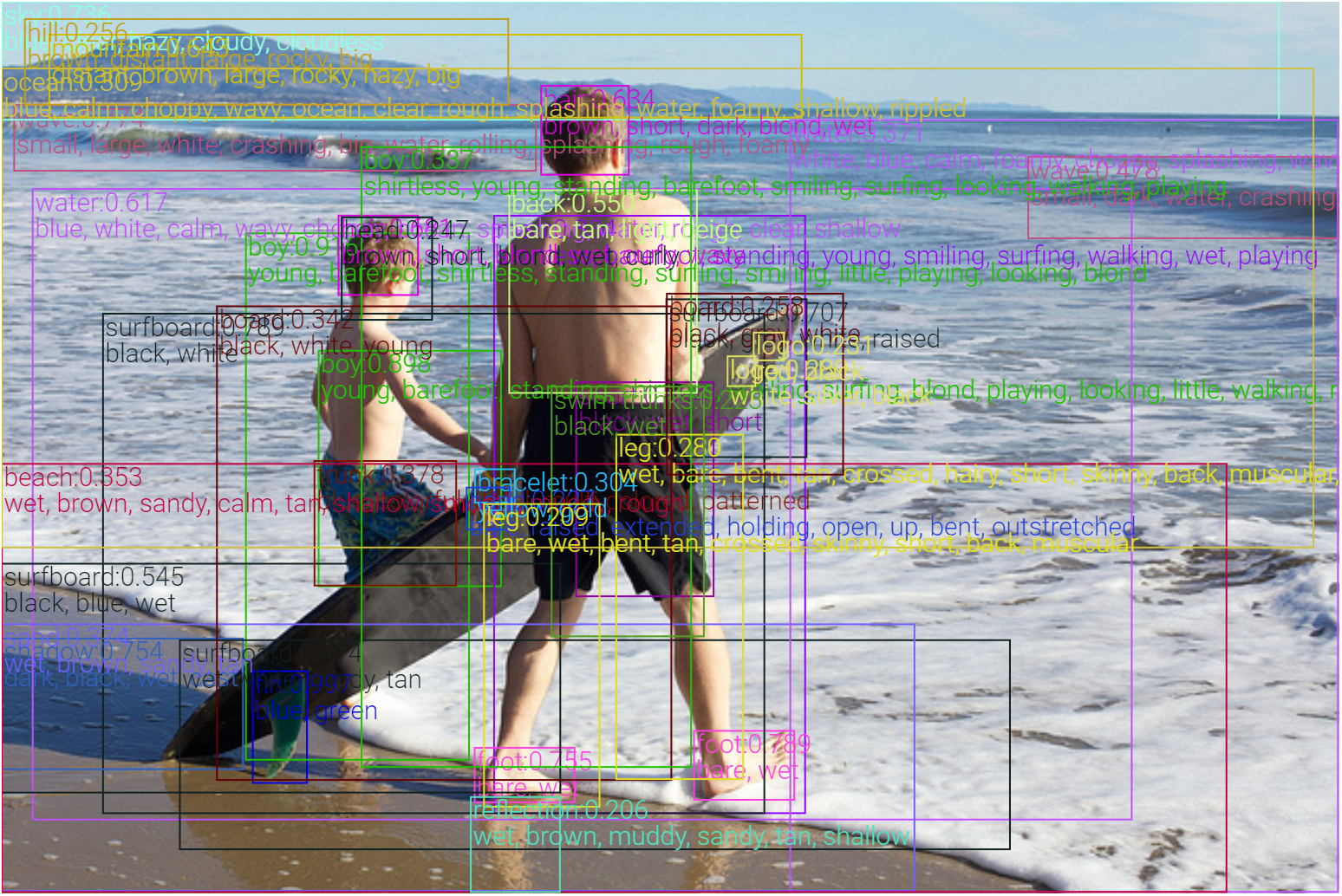}
\caption{Predictions from an X152-FPN model trained on OpenImages (Left) and our X152-C4 model trained on four public object detection datasets (Right). Our model contains much richer semantics, such as richer visual concepts and attribute information, and the detected bounding boxes cover nearly all semantically meaningful regions. Compared with those from the common object classes in typical OD models (Left), the rich and diverse region features from our model (Right) are crucial for vision-language tasks. For concepts detected by both models, e.g., ``\texttt{boy}'', attributes from our model offer richer information, e.g., ``\texttt{young barefoot shirtless standing surfing smiling little playing looking blond boy}''. There are object concepts that are detected by our model but not by the OpenImages model, including \texttt{fin}, \texttt{wave}, \texttt{foot}, \texttt{shadow}, \texttt{sky}, \texttt{hair}, \texttt{mountain}, \texttt{water}, \texttt{(bare}, \texttt{tan}, \texttt{light}, \texttt{beige}) \texttt{back}, (\texttt{blue}, \texttt{colorful}, \texttt{floral}, \texttt{multi colored}, \texttt{patterned}) \texttt{trunk}, \texttt{sand}, \texttt{beach}, \texttt{ocean}, (\texttt{yellow}, \texttt{gold}) \texttt{bracelet}, \texttt{logo}, \texttt{hill}, \texttt{head}, (\texttt{black}, \texttt{wet}) \texttt{swim trunks}, \texttt{black}, \texttt{wet swim trunks}. Compared to the R101-C4 model of \cite{anderson2018bottom}, our model produces more accurate object-attribute detection results and better visual features for VL applications; see Appendix~\ref{app:visionqa} for the full pictures and predictions from ~\cite{anderson2018bottom}.}
\label{fig:pretrain_finetune}
\end{figure*}


Among the aforementioned work, a widely-used object detection (OD) model~\cite{anderson2018bottom} is trained on the Visual Genome dataset~\cite{krishna2017visual}. The OD model provides an object-centric representation of images, and has been used in many VL models as a black box. In this work, we pre-train a large-scale object-attribute detection model based on the ResNeXt-152 C4 architecture (short as X152-C4).  
Compared to the OD model of ~\cite{anderson2018bottom}, the new model is better-designed for VL tasks, and is bigger and trained on much larger amounts of data, combining multiple public object detection datasets, including  COCO~\cite{lin2014microsoft}, OpenImages (OI)~\cite{kuznetsova2018open}, Objects365~\cite{shao2019objects365} and Visual Genome (VG)~\cite{krishna2017visual}. 
As a result, our OD model achieves much better results on a wide range of VL tasks, as shown in Table~\ref{tab:peter_vinvl_comp}.
Compared to other typical OD models, such as X152-FPN trained on OpenImages, our new model can encode a more diverse collection of visual objects and concepts (e.g., producing visual representations for $1848$ object categories and $524$ attribute categories), as illustrated by an example in Figure~\ref{fig:pretrain_finetune}. 

To validate the effectiveness of the new OD model, we pre-train a Transformer-based cross-modal fusion model \short~\cite{li2020oscar} on a public dataset consisting of $8.85$ million text-image pairs, where the visual representations of these images are produced by the new OD model and are fixed during \short{} pre-training. We then fine-tune the pre-trained \short{} for a wide range of downstream tasks, including VL understanding tasks such as VQA~\cite{goyal2017making}, GQA~\cite{hudson2019gqa}, NLVR2~\cite{suhr2018corpus}, and COCO text-image retrieval~\cite{lin2014microsoft}, and VL generation tasks such as COCO image captioning~\cite{lin2014microsoft} and NoCaps~\cite{agrawal2019nocaps}. Our results show that the object-centric representations produced by the new OD model significantly improve the performance across all the VL tasks, often by a large margin over strong baselines using the classical OD model~\cite{anderson2018bottom}, creating new state of the arts on all these tasks, including GQA on which none of the published pre-trained models has surpassed the deliberately designed neural state machine (NSM)~\cite{hudson2019learning}. We will release the new OD model to the research community.

The main contributions of this work can be summarized as follows:
$(\RN{1})$ We present a comprehensive empirical study to demonstrate that visual features matter in VL models.
$(\RN{2})$ We have developed a new object detection model that can produce better visual features of images than the classical OD model~\cite{anderson2018bottom} and substantially uplifts the state-of-the-art results on all major VL tasks across multiple public benchmarks. 
$(\RN{3})$ We provide a detailed ablation study of our pre-trained object detection model to investigate the relative contribution to the performance improvement due to different design choices regarding diversity of object categories, visual attribute training, training data scale, model size, and model architecture. 

\section{Improving Vision (V) in Vision Language (VL)}
\label{sec:vision_pretrain}
Deep learning-based VL models typically consist of two modules: an image understanding module $\mathbf{Vision}$ and a cross-modal understanding module $\mathbf{VL}$:
\begin{equation}\label{eq:vlm}
    (\qv, \vv) = \mathbf{Vision}(Img), \quad y = \mathbf{VL}(\wv, \qv, \vv),
\end{equation}
where $Img$ and $\wv$ are the inputs of the vision and language modalities, respectively. 
The output of the $\mathbf{Vision}$ module consists of $\qv$ and $\vv$.
$\qv$ is the semantic representation of the image, such as tags or detected objects, and $\vv$ the distributional representation of the image in a high-dimensional latent space represented using 
e.g., the box or region\footnote{We use the terms region and box interchangeably.} features produced by a VG-pre-trained Faster-RCNN model~\cite{anderson2018bottom}. 
Most $\mathbf{VL}$ models use only the visual features $\vv$, while the recently proposed \oscar~\cite{li2020oscar} model shows that $\qv$ can serve as anchors for learning better vision-language joint representations and and thus can improve the performance on various VL tasks. 
$\wv$ and $y$ of the $\mathbf{VL}$ module of Equation~\eqref{eq:vlm} vary among different VL tasks. In VQA, $\wv$ is a question and $y$ is an answer to be predicted. In text-image retrieval, $\wv$ is a sentence and $y$ is the matching score of a sentence-image pair. In image captioning, $\wv$ is not given and $y$ is a caption to be generated. 

Inspired by the great success of pre-trained language models to various natural language processing tasks, vision-language pre-training (VLP) has achieved remarkable success in improving the performance of the cross-modal understanding module $\mathbf{VL}$ by (1) unifying vision and language modeling $\mathbf{VL}$ with Transformer and (2) pre-training the unified $\mathbf{VL}$ with large-scale text-image corpora. However, most recent works on VLP treat the image understanding module $\mathbf{Vision}$ as a black box and leave the visual feature improvement untouched since the development of the classical OD model~\cite{anderson2018bottom} three years ago, despite that there has been much research progress on improving object detection by 1) developing much more diverse, richer, and larger training datasets (e.g. OpenImages and Objects 365), 2) gaining new insights in object detection algorithms such as feature pyramid network~\cite{lin2017feature}, one-stage dense prediction~\cite{lin2017focal}, and anchor-free detectors~\cite{tian2020fcos}, and 3) leveraging more powerful GPUs for training bigger models. 

In this work, we focus on improving $\mathbf{Vision}$ for better visual representations. We developed a new OD model by enriching the visual object and attribute categories, enlarging the model size and training on a much larger OD dasetset, and thus advanced the state of the arts on a wide range of VL tasks.
We detail how the new OD model is developed in the rest of this section and then describe the use of \short{} for $\mathbf{VL}$ pre-training in Section~\ref{sec:vl_pretrain}.

\subsection{Object Detection Pre-training}
\label{subsec:vision_pretrain}
To improve the OD model for VL tasks, we utilize four public object detection datasets. As most datasets do not have attribute annotations, we adopt a \emph{pre-training and fine-tuning} strategy to build our OD model. We first pre-train an OD model on a large-scale corpus consisting of four public datasets, and then fine-tune the model with an additional attribute branch on Visual Genome, making it capable of detecting both objects and attributes.

\paragraph{Data.} 
Table~\ref{tab:vision_pretrain_corpus} summarizes the statistics of the four public datasets used in our object detection pre-training, including COCO, OpenImagesV5 (OI), Objects365V1, and Visual Genome (VG). 
These datasets have complementary characters, and are extremely unbalanced in terms of data size, object vocabulary, and the number of annotations in each class. 
For example, the VG dataset has a rich and diverse set of annotations for both objects and their attributes with an open vocabulary. But its annotations are noisy and suffer from the missing-annotation problem. 
The COCO dataset, on the other hand, is very well annotated. But the coverage of visual objects and attributes is much lower than that in VG although we use both its 80 object classes and 91 stuff classes to include as diverse visual concepts as possible.
We take the following steps to build a unified corpus by combining the four datasets.
\begin{enumerate}
    \item First of all, to enhance visual concepts of tail classes, we perform class-aware sampling for OpenImages and Objects365 to get at least 2000 instances per class, resulting in 2.2M and 0.8M images, respectively. 
    \item To balance the contribution of each dataset, we merge the four datasets with 8 copies of COCO (8$\times$0.11M), 8 copies of VG (8$\times$0.1M), 2 copies of class-aware sampled Objects365 (2$\times$0.8M) and one copy of the class-aware sampled OpenImages (2.2M).
    \item To unify their object vocabularies, we use the VG vocabulary and its object aliases as the base vocabulary, merge a class from the other three datasets into a VG class if their class names or aliases match, and add a new class if no match is found.
    \item Finally, we keep all VG classes that contain at least 30 instances, resulting in 1594 VG classes and 254 classes from the other three datasets that cannot be mapped to the VG vocabulary, resulting in a merged object detection dataset that contains 1848 classes.
\end{enumerate}


\begin{table}[h]
\begin{center}
\begin{tabular}{c@{\hspace{3pt}}|c@{\hspace{3pt}}|c|c|c@{\hspace{3pt}}|c}
\toprule
Source & VG & COCO w/ stuff & Objects365 & OpenImagesV5 & Total \\
\midrule
Image & 97k & 111k & 609k & 1.67M & 2.49M  \\
classes & 1594 & 171 & 365 & 500 & 1848 \\
Sampling & $\times 8$ & $\times 8$ & CA-2k, $\times 2$ & CA-2k & 5.43M \\
\bottomrule
\end{tabular}
\end{center}
\vspace{-5mm}
\caption{Statistics of the Vision pre-training datasets. In sampling, $\times k$ means $k$ copies in one epoch and ``CA-2k'' means class-aware sampling with at least 2000 instances per class.}
\label{tab:vision_pretrain_corpus}
\vspace{-3mm}
\end{table}

\paragraph{Model Architecture (FPN vs C4).}  
Although \cite{lin2017feature} shows that the FPN model outperforms the C4 model for object detection, recent studies ~\cite{jiang2020defense} demonstrate that FPN does not provide more effective region features for VL tasks than C4, which is also confirmed by our experimental results
\footnote{We find in our experiments that using the same training process, the X152-C4 model even produces better object detection result than the X152-FPN model. See Appendix~\ref{app:c4vsfpn} for details.}. 
We thus conduct a set of carefully designed experiments, as to be detailed in Appendix~\ref{app:c4vsfpn}, and find two main reasons for this. 
The first is that all layers in the C4 model used for region feature extraction are pre-trained using the ImageNet dataset while the multi-layer-perceptron (MLP) head of the FPN model are not. It turns out that the VG dataset is still too small to train a good enough visual features for VL tasks and using ImageNet-pre-trained weights is beneficial.
The second is due to the different network architectures (CNN vs. MLP). The convolutional head used in C4 has a better inductive bias for encoding visual information than the MLP head of FPN. 
Therefore, in this study we use C4 architecture for VLP. 

\paragraph{Model Pre-Training.} Following the common practice in object detection training, we freeze the first convolution layer, the first residual block, and all the batch-norm layers. We also use several data augmentation methods, including horizontal flipping and multi-scale training. To train a detection model with the X152-C4 architecture, we initialize the model backbone from an ImageNet-5K checkpoint~\cite{wu2019detectron2} and train for 1.8M iterations with a batch size of 16 images.

\subsection{Injecting attribute information into the model}
Following \cite{anderson2018bottom}, we add an attribute branch to the pre-trained OD model, and then fine-tune the OD model on VG to inject attribute information (524 classes). 
Since the object representations are pre-trained in the object detection pre-training stage, we can focus the VG fine-tuning on learning attributes by picking a much larger attribute loss weight $1.25$, compared to $0.5$ used in \cite{anderson2018bottom,jiang2020defense}. 
Thus, our fine-tuned model significantly outperforms previous models \cite{anderson2018bottom,jiang2020defense} in detecting objects and attributes on VG. 

\subsection{Efficient region feature extractor for VL tasks}
With a richer set of visual objects and attributes, the classical class-aware non-maximal suppression (NMS) post-processing takes a significantly larger amount of time to remove overlapped bounding boxes, making the feature extraction process extremely slow. To improve the efficiency, we replace the class-aware NMS with the class-agnostic NMS that only conducts the NMS operation once\footnote{Counting the NMS in the RPN module, there are in total 2 NMS operations in our efficient region feature extractor.}. We also replace the time-consuming conv layers with dilation=2 used in \cite{anderson2018bottom} with conv layers without dilation. 
These two replacements make the region feature extraction process much faster than that in \cite{anderson2018bottom} without any accuracy drop on VL downstream tasks. We report the end-to-end inference time of VL models with different vision models on a Titan-X GPU and a CPU with a single thread in Table~\ref{tab:model_efficiency_cpu} in Appendix~\ref{appsec:gridfeature}. 

In summary, the pre-trained OD model serves as the image understanding module, 
as in Equation~\eqref{eq:vlm}, to produce vision presentations $(\qv, \vv)$ 
for downstream VL tasks. 
Here, $\qv$ is the set of detected object names (in text) and $\vv$ is the set of region features. 
Each region feature is denoted as $(\hat{v}, z)$, where $\hat{v}$ is a $P$-dimensional representation from the input of the last linear classification layer of the detection head (
\ie $P=2048$) and $z$ is a $R$-dimensional position encoding of the region (\ie $R=6$)\footnote{It includes coordinates of the bounding boxes, and height \& width.}. 

\section{\short{} Pre-training}
\label{sec:vl_pretrain}
The success of VLP lies in the use of a unifying model architecture for a wide range of VL tasks and the large-scale pre-training of the unified model using objectives that correlate with the performance metrics of these downstream VL tasks. 
In this study we pre-train an improved version of \oscar~ \cite{li2020oscar}, known as \short{} models, to learn the joint image-text representations using image tags as anchors for image-text alignment. 


\subsection{Pre-training corpus}
We build our pre-training corpus based on three types of existing vision and VL datasets: (1) image captioning datasets with human-annotated captions as $\wv$ and machine-generated~\footnote{We use the same model to extract visual features.} image tags as $\qv$, including COCO~\cite{lin2014microsoft}, Conceptual Captions (CC)~\cite{sharma2018conceptual}, SBU captions~\cite{ordonez2011im2text} and flicker30k~\cite{young2014image}; 
(2) visual QA datasets with questions as $\wv$ and human-annotated answers as $\qv$, including GQA~\cite{hudson2019gqa}, VQA~\cite{goyal2017making} and VG-QAs; 
(3) image tagging datasets with machine-generated~\footnote{We use the captioning model released by \oscar~\cite{li2020oscar}.} captions as $\wv$ and human-annotated tags as $\qv$, including a subset of OpenImages (1.67M images). 
In total, the corpus contains 5.65 million unique images, 8.85 million text-tag-image triples. The detailed statistics are presented in Table~\ref{tab:pretrain_corpus} in the Appendix. 
The size of the pre-training corpus could have been significantly increased by combining large-scale image tagging datasets, such as the full set of OpenImages (9M images) and YFCC (92M images). 
We leave it to future work to leverage much larger corpora for model pre-training. 

\begin{table*}[!ht]
\begin{center}
\begin{tabular}{c|cc|c|cc}
\toprule
Loss & \multicolumn{2}{c|}{$(\wv, \qv/\qv', \vv)$} & $(\wv/\wv', \qv, \vv)$ & \multicolumn{2}{c}{3-way contrastive} \\
$\wv'/\qv'$ & All $\qv$'s (\textsc{Oscar}) & $\qv$'s from QA & All $\wv$'s &  All (\short{}) & $\qv$'s from QA \\
\hline
VQA (vqa-dev) & {\bf 69.8}\small{$\pm$0.08} & \textcolor{blue}{\bf 70.1}\small{$\pm$0.08} &  69.5\small{$\pm$0.05} & {\bf 69.8}\small{$\pm$0.06} & {\bf 69.7}\small{$\pm$0.06} \\
COCO-IR & 73.9\small{$\pm$0.2} & {\bf 75.0}\small{$\pm$0.2} & {\bf 75.0}\small{$\pm$0.7} & \textcolor{blue}{\bf 78.3}\small{$\pm$0.3} & \textcolor{blue}{\bf 77.7}\small{$\pm$0.7}\\
\bottomrule
\end{tabular}
\caption{Effects of different pre-training contrastive losses on downstream tasks (R50-C4 as $\mathbf{Vision}$ module and 4-layer Transformer as $\mathbf{VL}$ module in \eqref{eq:vlm} ). COCO-IR metric is Image-to-Text retrieval R@1 at COCO 1K test set. \textcolor{blue}{\bf Blue} indicates the best result for a task and {\bf Black} indicates the runner-up. 
}
\label{tab:contrastive}
\end{center}
\vspace{-3mm}
\end{table*}

\subsection{Pre-training Objectives}
\label{subsec:oscarobjective}
There are two terms in the \short{} pre-training loss as in Equation~\eqref{eq_pre_training}.
\begin{align}
\Lcal_{\text{Pre-training}} = \Lcal_{\text{MTL}} + \Lcal_{\text{CL3}}.
\label{eq_pre_training}
\end{align}
$\Lcal_{\text{MTL}}$ is the {Masked Token Loss} defined on the text modality ($\wv$ and $\qv$), following closely~\cite{li2020oscar}. (See Appendix~\ref{app:oscarlosses} for details.)
$\Lcal_{\text{CL3}}$ is a novel \emph{3-way Contrastive Loss}.
Different from the binary contrastive loss used in \textsc{Oscar}~\cite{li2020oscar}, the proposed \emph{3-way Contrastive Loss} to effectively optimize the training objectives used for 
VQA \cite{yang2016stacked}
and text-image matching \cite{fang2015captions}\footnote{\cite{fang2015captions} uses a deep-learning-based text-image matching model to select the best caption candidate for a given image.}. 
As shown in Equation~\ref{eq_two_views}, $\Lcal_{\text{CL3}}$ takes into account two types of training samples $\xv$: the \{caption, image-tags, image-features\} triplets of the image captioning and image tagging data, and the \{question, answer, image-features\} triplets of the VQA data.

\begin{align} 
\xv 
\triangleq (\underbrace{~\wv_{~}}_{\text{\textcolor{red!50}{caption}} }, \underbrace{~\qv, \vv~}_{\text{\textcolor{mygreen}{tags\&image}}})
\quad \text{ or } \quad ( \underbrace{~\wv, \qv}_{\text{\textcolor{red!50}{Q\&A}}},~ \underbrace{~\vv_{~_{~}}}_{\text{\textcolor{mygreen}{image}}})  
\label{eq_two_views}
\end{align}

To compute contrastive losses, negative examples need to be constructed. 
We construct two types of negative (unmatched) triplets for the two types of training samples, respectively. 
One is the polluted ``captions'' $(\wv', \qv, \vv)$ and the other the polluted ``answers'' $(\wv, \qv', \vv)$. 
To classify whether a caption-tags-image triplet contains a polluted caption is a text-image matching task. 
To classify whether a question-answer-image triplet contains a polluted answer is an answer selection task for VQA. 
Since the encoding of $\mathtt{[CLS]}$ can be viewed as a representation of 
the triplet $(\wv, \qv, \vv)$, we apply a fully-connected (FC) layer on top of it as a 3-way classifier $f(.)$ to predict whether the triplet is matched ($c=0$), contains a polluted $\wv$ ($c=1$), or contains a polluted $\qv$ ($c=2$). The 3-way contrastive loss is defined as
\begin{align}
\Lcal_{\text{CL3}} = -\E_{  (\wv, \qv, \vv; c ) \sim \Tilde{\Dcal} } \log p( c | f(\wv, \qv, \vv) ),
\label{eq_action_prediction}
\end{align}
where the dataset $(\wv, \qv, \vv; c ) \in \Tilde{\Dcal}$ contains  50\% matched triples, 25\% $\wv$-polluted triples, and 25\% $\qv$-polluted triples. 
For efficient implementation, the polluted $\wv'$ is uniformly sampled from all $\wv$'s (captions and questions) and $\qv'$ is uniformly sampled from all $\qv$'s (tags and answers) in the corpus. 
As demonstrated in Table~\ref{tab:contrastive}, when only the answer-polluted triplets are used, i.e.,  $(\wv, \qv', \vv)$ with $\qv'$ sampled from $\qv$'s from QA corpus, the contrastive loss simulates closely the objective for the VQA task but not the text-image retrieval task. As a result, the pre-trained model can be effectively adapted to VQA, but not so to text-image retrieval. 
By contrast, the proposed 3-way contrastive loss transfers well to both tasks.


\subsection{Pre-trained models} 
We pre-train two model variants, denoted as \shortb{} and \shortl{}, which are initialized with parameters $\thetav_{\text{BERT}}$ of BERT base ($L=12, H=768, A=12$) and large ($L=24, H=1024, A=16$), respectively, where $L$ is the number of layers, $H$ the hidden size, and $A$ the number of self-attention heads. 
To ensure that the image region features have the same input embedding size as BERT, we transform the position-augmented region features using a linear projection via matrix $\Wmat$. The trainable parameters are $\thetav=\{\thetav_{\text{BERT}}, \Wmat\}$. 
\shortb{} is trained for at least $1$M steps, with learning rate $1e^{-4}$ and batch size $1024$. \shortl{} is trained for at least $1$M steps, with learning rate $3e^{-5}$ and batch size $1024$. The sequence length of language tokens $[\wv, \qv]$ and region features $\vv$ are $35$ and $50$, respectively.




\section{Adapting to VL Tasks} 
\label{sec:downstream_tasks}
%
We adapt the pre-trained models to seven downstream VL tasks, including five understanding tasks and two generation tasks. Each task poses different challenges for adaptation. This section briefly introduces the tasks and our fine-tuning strategy. We refer the readers to Appendix~\ref{sec:downstreams} for details.

\paragraph{VQA \& GQA} These two are the most widely used understanding task for evaluating VL models in the research community. 
The tasks require the model to answer natural language questions based on an image. 
In this study, we perform experiments on the widely-used VQA v2.0 dataset~\cite{goyal2017making} and GQA dataset~\cite{hudson2019gqa}, 
Following the setting of~\cite{anderson2018bottom}, for each question, the model picks an answer from a shared answer set (i.e., $3,129$ candidates for VQA, $1,852$ candidates for GQA).
When adapting a VLP model to the VQA task, we construct the input by concatenating a given question, object tags and object region features, and then feed the $\mathtt{[CLS]}$ output from \short{} to a task-specific linear classifier with a softmax layer for answer prediction.

\paragraph{Image Captioning \& NoCaps} The captioning task is to generate a natural language caption for an image. This is the most widely used VL generation task in the research community -- the Image Captioning Leaderboard
\footnote{Image Captioning Leaderboard: \url{https://competitions.codalab.org/competitions/3221}}
hosts more than 260 models as of December 10, 2020. 
To enable caption generation, we fine-tune \short{} using the seq2seq objective. Each training sample is converted to a triplet consisting of a caption, a set of image region features, and a set of object tags. 
We randomly mask out $15\%$ of the caption tokens, and use the encoding of the remaining context (the triplet) 
to predict the masked tokens.
Similar to VLP~\cite{li2020oscar,zhou2019unified}, the self-attention mask is constrained such that a caption token can only attend to the tokens before its position to simulate a uni-directional generation process.  
All caption tokens have full attentions to image regions and object tags but not the other way around. 
During inference, we first encode the image regions, object tags, and a special token $\mathtt{[CLS]}$ as input. Then the model starts to generate a caption by feeding in a $\mathtt{[MASK]}$ token and sampling a token from a vocabulary based on the token probability output. Next, the $\mathtt{[MASK]}$ token in the previous input sequence is replaced with the sampled token and a new $\mathtt{[MASK]}$ is appended for the next word prediction. The generation process terminates when the model outputs the $\mathtt{[STOP]}$ token or the generated sentence exceeds a pre-defined max length. 
We perform image captioning experiments on the COCO image captioning dataset~\cite{lin2014microsoft}. \textbf{N}ovel \textbf{O}bject \textbf{Cap}tioning at \textbf{S}cale~\cite{agrawal2019nocaps} extends the image captioning task to test a model's capability of describing novel objects from the Open Images dataset~\cite{kuznetsova2018open} which are unseen in the training corpus. Following the restriction guideline of NoCaps, we use the predicted Visual Genome and Open Images labels to form the input tag sequences, and directly train \short{} on COCO without the initialization from pre-training. VIVO~\cite{hu2020vivo} proposed a VLP technique by only using image tagging data, and achieved SOTA results on NoCaps by fine-tuning on COCO captions. We reproduced VIVO with only one change, i.e., replacing its original vision model with our new vision model, and improved the VIVO performance significantly (short as VinVL+VIVO), as reported in Table~\ref{tab:nocaps_caption}.

\paragraph{Image(-to-Text) Retrieval \& Text(-to-Image) Retrieval} 
Both tasks require the model to calculate a similarity score between an image and a sentence. Thus, the task is widely used to directly measure the quality of the cross-modal VL representation.
%
Following~\cite{li2020oscar}, we formulate the task as a binary classification problem, where given a matched image-text pair, we randomly select a different image or a different sentence to form an unmatched pair. The representation of $\mathtt{[CLS]}$ is used as the input to a classifier to predict a score indicating how likely the given pair is matched. 
In testing, the predicted score is used to rank a given image-text pairs of a query. Following~\cite{li2019unicoder}, we report the top-$K$ retrieval results on both the $1$K and $5$K COCO test sets.

\paragraph{NLVR2} The dataset is developed for joint reasoning about natural language and images~\cite{suhr2018corpus}. 
The task is to determine whether a text description is true about a pair of images.
%
For fine-tuning, we first construct two input sequences, each containing the concatenation of the given text description and one of the images, and then two $\mathtt{[CLS]}$ outputs from \short{} are concatenated to form the input to a binary classifier for prediction.

\section{Experiments \& Analysis}
\label{sec:exps}

\subsection{Main Results}
\label{subsec:sotas}
To account for model parameter efficiency, we group the SoTA models in three categories: 
$(\RN{1})$ SoTA$_{S}$ indicates the best performance achieved by small models prior to the Transformer-based VLP models.
$(\RN{2})$ SoTA$_{B}$ indicates the best performance produced by VLP models of a similar size to BERT base. 
$(\RN{3})$ SoTA$_{L}$ indicates the best performance yielded by VLP models that have a similar size to BERT large.

Table~\ref{tab:overall_result}
gives an overview of the results of \short{} with \vinvl (short for \vinvl) on seven VL tasks, 
compared to previous SoTAs\footnote{All the (single-model) SoTAs are from the published results. For all the tables in this paper, \textcolor{blue}{\textbf{Blue}} indicates the best result for a task, and gray background indicates results produced by \vinvl.}.
\vinvl outperforms previous SoTA models on all tasks\footnote{The only exception is B@4 on image captioning.}, 
often by a significantly large margin. The result demonstrates the effectiveness of the region features produced by the new OD model.

\begin{table*}[!ht]
\begin{center}
\resizebox{\linewidth}{!}{
\scriptsize
\begin{tabular}{c@{\hspace{3pt}}|c@{\hspace{5pt}}c@{\hspace{3pt}}|c@{\hspace{5pt}}c|c@{\hspace{5pt}}c@{\hspace{5pt}}c@{\hspace{6pt}}c@{\hspace{5pt}}|c@{\hspace{6pt}}c@{\hspace{6pt}}|c@{\hspace{6pt}}c@{\hspace{5pt}}c@{\hspace{6pt}}|c@{\hspace{6pt}}c@{\hspace{6pt}}c@{\hspace{3pt}}|c@{\hspace{6pt}}c}
\toprule
\multirow{2}{*}{Task} & \multicolumn{2}{c|}{VQA} & \multicolumn{2}{c|}{GQA} & \multicolumn{4}{c|}{Image Captioning} & \multicolumn{2}{c|}{NoCaps} & \multicolumn{3}{c|}{Image Retrieval} & \multicolumn{3}{c|}{Text Retrieval} & \multicolumn{2}{c}{NLVR2} \\ 
& test-dev & test-std & test-dev & test-std & B@4 & M & C & S & C & S & R@1 & R@5 & R@10 & R@1 & R@5 & R@10 & dev & test-P \\ \midrule
SoTA$_{S}$ & $70.55$ & $70.92$ & $-$ & $\bf 63.17$ & $38.9$ & $29.2$ & $129.8$ & $22.4$ & $61.5$ & $9.2$ & $39.2$ & $68.0$ & $81.3$ & $56.6$ & $84.5$ & $92.0$ & $54.10$ & $54.80$ \\
SoTA$_{B}$ & $73.59$ & $73.67$ & $61.58$ & $61.62$ & $40.5$ & $29.7$ & $137.6$ & $22.8$ & $86.58$ & $12.38$ & $54.0$ & $80.8$ & $88.5$ & $70.0$ & $91.1$ & $95.5$ & $78.39$ & $79.30$ \\
SoTA$_{L}$ & $74.75$ & $74.93$ & $-$ & $-$ & \textcolor{blue}{$\bf 41.7$} & $30.6$ & $140.0$ & $24.5$ & $-$ & $-$ & $57.5$ & $82.8$ & $89.8$ & $73.5$ & $92.3$ & $96.0$ & $79.76$ & $81.47$ \\
\hline
\rowcolor{Graylight}  
\vinvlb{} & $\bf 75.95$ & $\bf 76.12$ & \textcolor{blue}{$\bf 65.05$} & \textcolor{blue}{$\bf 64.65$} & $40.9$ & $\bf 30.9$ & \textcolor{blue}{$\bf 140.6$} & $\bf 25.1$ & $\textcolor{blue}{\bf 92.46}$ & \textcolor{blue}{$\bf 13.07$} & $\bf 58.1$ & $\bf 83.2$ & $\bf 90.1$ & $\bf 74.6$ & $\bf 92.6$ & \textcolor{blue}{$\bf 96.3$} & $\bf 82.05$ & $\bf 83.08$ \\ 
\rowcolor{Grayheavy}
\vinvll{} & \textcolor{blue}{$\bf 76.52$} & \textcolor{blue}{$\bf 76.60$} & $-$ & $-$ & $\bf 41.0$ & \textcolor{blue}{$\bf 31.1$} & \textcolor{blue}{$\bf 140.9$} & \textcolor{blue}{$\bf 25.2$} & $-$ & $-$ & \textcolor{blue}{$\bf 58.8$} & \textcolor{blue}{$\bf 83.5$} & \textcolor{blue}{$\bf 90.3$} & \textcolor{blue}{$\bf 75.4$} & \textcolor{blue}{$\bf 92.9$} & $\bf 96.2$ & \textcolor{blue}{$\bf 82.67$} & \textcolor{blue}{$\bf 83.98$} \\
\hline
$\Delta$ & $\bf 1.77\uparrow$ & $\bf 1.67\uparrow$ & $\bf 3.47\uparrow$ & $\bf 1.48\uparrow$ & $0.7\downarrow$ & $\bf 0.5\uparrow$ & $\bf 0.9\uparrow$ & $\bf 0.7\uparrow$ & $\bf 5.9\uparrow$ & $\bf 0.7\uparrow$ & $\bf 1.3\uparrow$ & $\bf 0.7\uparrow$ & $\bf 0.5\uparrow$ & $\bf 1.9\uparrow$ & $\bf 0.6\uparrow$ & $\bf 0.3\uparrow$ & $\bf 2.91\uparrow$ & $\bf 2.51\uparrow$ \\ 
\bottomrule
\end{tabular}
}
\end{center}
\vspace{-5mm}
\caption{
An overall comparison with SoTAs on seven tasks. $\Delta$ indicates the improvement over SoTA. SoTA with subscript S, B, L indicates performance achieved by small models, and models with the model size similar to BERT base and large, respectively. SoTAs: VQA is from ERNIE-VIL~\cite{yu2020ernie}, GQA is from NSM~\cite{hudson2019learning}, NoCaps is from VIVO~\cite{hu2020vivo}, NLVR2 is from VILLA~\cite{gan2020large}, the rest tasks are from \textsc{Oscar}~\cite{li2020oscar}.}
\label{tab:overall_result}
\vspace{-3mm}
\end{table*}

\begin{table*}[ht]
\begin{center}
\resizebox{\linewidth}{!}{
\scriptsize
\begin{tabular}{l@{\hspace{2pt}}|c@{\hspace{1pt}}c@{\hspace{1pt}}c@{\hspace{1pt}}c@{\hspace{1pt}}c@{\hspace{3pt}}c@{\hspace{5pt}}c@{\hspace{5pt}}c@{\hspace{5pt}}c@{\hspace{5pt}}c@{\hspace{5pt}}c@{\hspace{5pt}}c@{\hspace{5pt}}c@{\hspace{3pt}}|c@{\hspace{3pt}}|>{\columncolor[gray]{0.90}}c@{\hspace{6pt}}>{\columncolor[gray]{0.90}}c}
\toprule
\multirow{2}{*}{Method} & \multicolumn{1}{c}{ViLBERT} & \multicolumn{1}{c}{VL-BERT} & \multicolumn{1}{c}{VisualBERT} & \multicolumn{1}{c}{LXMERT} & \multicolumn{1}{c}{12-in-1} & \multicolumn{2}{c}{\textsc{UNITER}} & \multicolumn{2}{c}{\textsc{Oscar}} & \multicolumn{2}{c}{\textsc{VILLA}} & \multicolumn{2}{c|}{\textsc{ERNIE-ViL}} & InterBERT & \multicolumn{2}{c}{\short w/ \vinvl} \\ 
& Base & Base & Base & Base & Base & Base & Large & Base & Large & Base & Large & Base & Large & \textcolor{blue}{\textbf{Ensemble}*} & Base & Large \\ \midrule
Test-dev & $70.63$ & $70.50$ & $70.80$ & $72.42$ & $73.15$ & $72.27$ & $73.24$ & $73.16$ & $73.61$ & $73.59$ & $73.69$ & $72.62$ & $74.75$ & - & $\bf 75.95$ & \textcolor{blue}{$\bf 76.52$} \\
Test-std & $70.92$ & $70.83$ & $71.00$ & $72.54$ & $-$ & $72.46$ & $73.40$ & $73.44$ & $73.82$ & $73.67$ & $74.87$ & $72.85$ & $74.93$ & $76.10$ & $\bf 76.12$ & \textcolor{blue}{$\bf 76.60$} \\
\bottomrule
\end{tabular}}
\end{center}
\vspace{-5mm}
\caption{Evaluation results on VQA. \textcolor{blue}{*} denotes the No.1 ensemble model of InterBERT Large on the VQA leaderboard.}
\label{tab:vqa_result}
\vspace{-3mm}
\end{table*}

\begin{table}[ht]
\begin{center}
\scriptsize
\begin{tabular}{l|c@{\hspace{5pt}}c@{\hspace{5pt}}c@{\hspace{5pt}}c@{\hspace{5pt}}c|>{\columncolor[gray]{0.90}}c>{\columncolor[gray]{0.90}}c}
\toprule
Method & LXMERT & MMN~\cite{chen2019meta} & 12-in-1 & \textsc{Oscar}$_\text{B}$ & NSM~\cite{hudson2019learning} & \shortb{} w/ \vinvl  \\ \midrule
Test-dev & $60.00$ & $-$ & $-$ & $\bf 61.58$ & $-$ & \textcolor{blue}{$\bf 65.05$} \\
Test-std & $60.33$ &  $60.83$ & $60.65$ & $61.62$ & $\bf 63.17$ & \textcolor{blue}{$\bf 64.65$} \\
\bottomrule
\end{tabular}
\end{center}
\vspace{-5mm}
\caption{Evaluation results on GQA.}
\label{tab:gqa_result}
\vspace{-3mm}
\end{table}

\begin{table}[ht]
\begin{center}
\scriptsize
\begin{tabular}{l@{\hspace{3pt}}|c@{\hspace{6pt}}c@{\hspace{6pt}}c@{\hspace{6pt}}c@{\hspace{5pt}}|c@{\hspace{6pt}}c@{\hspace{6pt}}c@{\hspace{6pt}}c}
\toprule
\multirow{2}{*}{Method} & \multicolumn{4}{c|}{cross-entropy optimization} & \multicolumn{4}{c}{CIDEr optimization}\\ 
& B@4 & M & C & S & B@4 & M & C & S\\ \midrule
BUTD~\cite{anderson2018bottom} & $36.2$ & $27.0$ & $113.5$ & $20.3$ & $36.3$ & $27.7$ & $120.1$ & $21.4$ \\
VLP~\cite{zhou2019unified} & $36.5$ & $28.4$ & $117.7$ & $21.3$ & $39.5$ & $29.3$ & $129.3$ & $23.2$ \\ 
AoANet~\cite{huang2019attention} & $37.2$ & $28.4$ & $119.8$ & $21.3$ & $38.9$ & $29.2$ & $129.8$ & $22.4$ \\ 
\textsc{Oscar}$_\text{B}$~\cite{li2020oscar} & $36.5$ & $30.3$ & $123.7$ & $23.1$ & $40.5$ & $29.7$ & $137.6$ & $22.8$ \\
\textsc{Oscar}$_\text{L}$~\cite{li2020oscar} & $37.4$ & \textcolor{blue}{$\bf 30.7$} & $127.8$ & $\bf 23.5$ & \textcolor{blue}{$\bf 41.7$} & {$30.6$} & $140.0$ & $24.5$ \\ \hline
\rowcolor{Gray} 
\shortb{} w/ \vinvl & $\bf 38.2$ & $30.3$ & \textcolor{blue}{$\bf 129.3$} & \textcolor{blue}{$\bf 23.6$} & $40.9$ & $\bf 30.9$ & $\bf 140.4$ & $\bf 25.1$ \\
\rowcolor{Gray}
\shortl{} w/ \vinvl & \textcolor{blue}{$\bf 38.5$} & $\bf 30.4$ & \textcolor{blue}{$\bf 130.8$} & $23.4$ & $\bf 41.0$ & \textcolor{blue}{$\bf 31.1$} & \textcolor{blue}{$\bf 140.9$} & \textcolor{blue}{$\bf 25.2$} \\
\bottomrule
\end{tabular}
\end{center}
\vspace{-5mm}
\caption{Image captioning evaluation results (single model) on COCO ``Karpathy" test split. (Note: B@4: BLEU@4, M: METEOR, C: CIDEr, S: SPICE.)}
\label{tab:mscoco_caption}
\vspace{-2mm}
\end{table}

\begin{table*}[ht]
\begin{center}
\scriptsize
\begin{tabular}{l@{\hspace{3pt}}|c@{\hspace{6pt}}c@{\hspace{6pt}}|c@{\hspace{6pt}}c@{\hspace{6pt}}|c@{\hspace{6pt}}c@{\hspace{6pt}}|c@{\hspace{6pt}}c@{\hspace{6pt}}|c@{\hspace{6pt}}c@{\hspace{6pt}}|c@{\hspace{6pt}}c@{\hspace{6pt}}|c@{\hspace{6pt}}c}
\toprule
\multirow{2}{*}{Method} & \multicolumn{2}{c|}{BLEU@1} & \multicolumn{2}{c|}{BLEU@2} & \multicolumn{2}{c|}{BLEU@3} & \multicolumn{2}{c|}{BLEU@4} & \multicolumn{2}{c|}{METEOR} & \multicolumn{2}{c|}{ROUGE-L} & \multicolumn{2}{c}{CIDEr-D} \\ 
& c5 & c40 & c5 & c40 & c5 & c40 & c5 & c40 & c5 & c40 & c5 & c40 & c5 & c40 \\ \midrule
BUTD~\cite{anderson2018bottom} & $80.2$ & $95.2$ & $64.1$ & $88.8$ & $49.1$ & $79.4$ & $36.9$ & $68.5$ & $27.6$ & $36.7$ & $57.1$ & $72.4$ & $117.9$ & $120.5$ \\
AoANet~\cite{huang2019attention} & $81.0$ & $95.0$ & $65.8$ & $89.6$ & $51.4$ & $81.3$ & $39.4$ & $71.2$ & $29.1$ & $38.5$ & $58.9$ & $74.5$ & $126.9$ & $129.6$ \\
X-Transformer~\cite{pan2020x} & $81.9$ & $95.7$ & $66.9$ & $90.5$ & $52.4$ & $82.5$ & $40.3$ & $72.4$ & $29.6$ & $39.2$ & $59.5$ & $75.0$ & $131.1$ & $133.5$ \\ 
\hline
\rowcolor{Gray} 
\short{} w/ \vinvl & \textcolor{blue}{$\bf 81.9$} & \textcolor{blue}{$\bf 96.9$} & \textcolor{blue}{$\bf 66.9$} & \textcolor{blue}{$\bf 92.4$} & \textcolor{blue}{$\bf  52.6$} & \textcolor{blue}{$\bf 84.7$} & \textcolor{blue}{$\bf 40.4$} & \textcolor{blue}{$\bf 74.9$} & \textcolor{blue}{$\bf 30.6$} & \textcolor{blue}{$\bf 40.8$} & \textcolor{blue}{$\bf 60.4$} & \textcolor{blue}{$\bf 76.8$} & \textcolor{blue}{$\bf 134.7$} & \textcolor{blue}{$\bf 138.7$} \\
\bottomrule
\end{tabular}
\end{center}
\vspace{-5mm}
\caption{Leaderboard of the state-of-the-art image captioning models on the COCO online testing.}
\label{tab:coco_leaderboard}
\vspace{-2mm}
\end{table*}

\begin{table}[h!]
\begin{center}
{\fontsize{7}{8}\selectfont 
\begin{tabular}{l@{\hspace{3pt}}@{\hspace{3pt}}|c@{\hspace{5pt}}c@{\hspace{1mm}}|c@{\hspace{5pt}}c@{\hspace{1mm}}|c@{\hspace{5pt}}c@{\hspace{1mm}}|c@{\hspace{5pt}}c@{\hspace{1pt}}@{\hspace{2pt}}|c@{\hspace{5pt}}c@{\hspace{1mm}}|c@{\hspace{5pt}}c@{\hspace{1mm}}|c@{\hspace{5pt}}c@{\hspace{1mm}}|c@{\hspace{5pt}}c}
\toprule
\multirow{2}{*}{Method } & \multicolumn{2}{c|}{in-domain} & \multicolumn{2}{c|}{near-domain} & \multicolumn{2}{c|}{out-of-domain} & \multicolumn{2}{c|}{overall} & \multicolumn{2}{c|}{in-domain} & \multicolumn{2}{c|}{near-domain} & \multicolumn{2}{c|}{out-of-domain} & \multicolumn{2}{c}{overall} \\ 
& CIDEr & SPICE & CIDEr & SPICE & CIDEr & SPICE & CIDEr & SPICE & CIDEr & SPICE & CIDEr & SPICE & CIDEr & SPICE & CIDEr & SPICE \\ \midrule
 \multicolumn{9}{c|}{Validation Set} & \multicolumn{8}{c}{Test Set} \\ \hline
UpDown$^+$ & $79.3$ & $12.4$ & $73.8$ & $11.4$ & $71.7$ & $9.9$ & $74.3$ & $11.2$ & $76.0$ & $11.8$ & $74.2$ & $11.5$ & $66.7$ & $9.7$ & $73.1$ & $11.2$ \\
\oscar$_\text{B}$* & $83.4$ & $12.0$ & $81.6$ & $12.0$ & $77.6$ & $10.6$ & $81.1$ & $11.7$ & $81.3$ & $11.9$ & $79.6$ & $11.9$ & $73.6$ & $10.6$ & $78.8$ & $11.7$ \\ 
\oscar$_\text{L}$* & $85.4$ & $11.9$ & $84.0$ & $11.7$ & $80.3$ & $10.0$ & $83.4$ & $11.4$ & $84.8$ & $12.1$ & $82.1$ & $11.5$ & $73.8$ & $9.7$ & $80.9$ & $11.3$ \\ 
Human~\cite{agrawal2019nocaps} & $84.4$ & \textcolor{blue}{$\bf 14.3$} & $ 85.0$ & \textcolor{blue}{$\bf 14.3$} & \textcolor{blue}{$\bf 95.7$} & \textcolor{blue}{$\bf 14.0$} & $87.1$ & \textcolor{blue}{$\bf 14.2$} & $80.6$ & \textcolor{blue}{$\bf 15.0$} & $84.6$ & \textcolor{blue}{$\bf 14.7$} & \textcolor{blue}{$\bf 91.6$} & \textcolor{blue}{$\bf 14.2$} & $85.3$ & \textcolor{blue}{$\bf 14.6$} \\ 
VIVO*~\cite{hu2020vivo} & $92.2$ & $12.9$ & $87.8$ & $12.6$ & $\bf 87.5$ & $11.5$ & $88.3$ & $12.4$ & $89.0$ & $12.9$ & $87.8$ & $12.6$ & $\bf 80.1$ & $11.1$ & $\bf 86.6$ & $12.4$ \\
\midrule
\rowcolor{Gray} 
VinVL* & {$\bf 96.8$} & $13.5$ & {$\bf 90.7$} & $13.1$ & $87.4$ & {$11.6$} & {$\bf 90.9$} & $12.8$ & {$\bf 93.8$} & $13.3$ & {$\bf 89.0$} & $12.8$ & $66.1$ & $10.9$ & {$85.5$} & $12.5$ \\ 
\rowcolor{Gray}
VinVL+VIVO & \textcolor{blue}{$\bf 103.7$} & $\bf 13.7$ & \textcolor{blue}{$\bf 95.6$} & $\bf 13.4$ & $83.8$ & $\bf 11.9$ & \textcolor{blue}{$\bf 94.3$} & $\bf 13.1$ & \textcolor{blue}{$\bf 98.0$} & $\bf 13.6$ & \textcolor{blue}{$\bf 95.2$} & $\bf 13.4$ & $78.0$ & $\bf 11.5$ & \textcolor{blue}{$\bf 92.5$} & $\bf 13.1$ \\ 
\bottomrule
\end{tabular}
}
\end{center}
\vspace{-5mm}
\caption{NoCaps evaluation results.
All the models are trained on COCO without additional image-caption pairs following the restriction of NoCaps.
(UpDown$^+$ is UpDown+ELMo+CBS, the models with * is +SCST+CBS, VinVL+VIVO is with SCST only.)}
\label{tab:nocaps_caption}
\vspace{-3mm}
\end{table}

\begin{table}[t!]
\begin{center}
\scriptsize
\begin{tabular}{@{\hspace{1mm}}l@{\hspace{1mm}}c@{\hspace{2mm}}c@{\hspace{2mm}}c@{\hspace{2mm}}c@{\hspace{2mm}}|c@{\hspace{2mm}}c@{\hspace{2mm}}c@{\hspace{1mm}}@{\hspace{2mm}}|c@{\hspace{2mm}}c@{\hspace{2mm}}c@{\hspace{2mm}}|c@{\hspace{2mm}}c@{\hspace{2mm}}c@{\hspace{1mm}}}
\toprule
\multirow{3}{*}{Method $\downarrow$} & \multirow{3}{*}{\texttt{BERT}} & \multicolumn{6}{c}{1K Test Set} & \multicolumn{6}{c}{5K Test Set} \\
\cline{3-14}
& & \multicolumn{3}{c|}{Text Retrieval} & \multicolumn{3}{c|}{Image Retrieval} & \multicolumn{3}{c|}{Text Retrieval} & \multicolumn{3}{c}{Image Retrieval} \\
& & R@1 & R@5 & R@10 & R@1 & R@5 & R@10 & R@1 & R@5 & R@10 & R@1 & R@5 & R@10 \\ 

\midrule
Unicoder-VL~\cite{li2019unicoder} & \texttt{B} & $84.3$ & $97.3$ & $99.3$ & $69.7$ & $93.5$ & $97.2$ & $62.3$ & $87.1$ & $92.8$ & $46.7$ & $76.0$ & $85.3$ \\ 
\multirow{2}{*}{\textsc{UNITER}~\cite{chen2019uniter}} & \texttt{B} & $-$ & $-$ & $-$ & $-$ & $-$ & $-$ & $63.3$ & $87.0$ & $93.1$ & $48.4$ & $76.7$ & $85.9$ \\
 & \texttt{L} & $-$ & $-$ & $-$ & $-$ & $-$ & $-$ & $66.6$ & $89.4$ & $94.3$ & $51.7$ & $78.4$ & $86.9$ \\
\multirow{2}{*}{\textsc{Oscar}} & \texttt{B} & $88.4$ & $99.1$ & $99.8$ & $75.7$ & $95.2$ & $98.3$ & $70.0$ & $91.1$ & $95.5$ & $54.0$ & $80.8$ & $88.5$ \\ 
 & \texttt{L} & $89.8$ & {$\bf 98.8$} & $\bf 99.7$ & $\bf 78.2$ & $\bf 95.8$ & $\bf 98.3$ & $73.5$ & $92.2$ & $96.0$ & $57.5$ & $82.8$ & $89.8$ \\ \hline 
\multirow{2}{*}{\short{} w/ \vinvl} & \texttt{B} & $\bf 89.8$ & {$\bf 98.8$} & $\bf 99.7$ & $\bf 78.2$ & $95.6$ & $98.0$ & $\bf 74.6$ & {$\bf 92.6$} & \textcolor{blue}{$\bf 96.3$} & $\bf 58.1$ & $\bf 83.2$ & $\bf 90.1$ \\
 & \texttt{L} & \textcolor{blue}{$\bf 90.8$} & \textcolor{blue}{$\bf 99.0$} & \textcolor{blue}{$\bf 99.8$} & \textcolor{blue}{$\bf 78.8$} & \textcolor{blue}{$\bf 96.1$} & \textcolor{blue}{$\bf 98.5$} & \textcolor{blue}{$\bf 75.4$} & \textcolor{blue}{$\bf 92.9$} & $\bf 96.2$ & \textcolor{blue}{$\bf 58.8$} & \textcolor{blue}{$\bf 83.5$} & \textcolor{blue}{$\bf 90.3$} \\
\bottomrule
\end{tabular}
\end{center}
\vspace{-5mm}
\caption{Text and Image retrieval evaluation on the COCO $1$K and $5$K test sets. (\texttt{B} for Base, \texttt{L} for Large)}
\label{tab:mscoco_retrieval}
\end{table}


\begin{table*}[ht!]
\begin{center}
\scriptsize
\begin{tabular}{l|cccccccccc|>{\columncolor[gray]{0.90}}c>{\columncolor[gray]{0.90}}c}
\toprule
\multirow{2}{*}{Method} & \multicolumn{1}{c}{MAC} & \multicolumn{1}{c}{VisualBERT} & \multicolumn{1}{c}{LXMERT} & \multicolumn{1}{c}{12-in-1} & \multicolumn{2}{c}{\textsc{UNITER}} & \multicolumn{2}{c}{\textsc{Oscar}} & \multicolumn{2}{c|}{\textsc{VILLA}} &  \multicolumn{2}{c}{\short w/ \vinvl}\\ 
& & base & base & base & base & large & base & large & base & large & base & large \\\midrule
Dev & $50.8$ & $67.40$ & $74.90$ & $-$ & $77.14$ & $78.40$ & $78.07$ & $79.12$ & $78.39$ & $79.76$ & $\bf 82.05$ & \textcolor{blue}{$\bf 82.67$} \\
Test-P & $51.4$ & $67.00$ & $74.50$ & $78.87$ & $77.87$ & $79.50$ & $78.36$ & $80.37$ & $79.47$ & $81.47$ & $\bf 83.08$ & \textcolor{blue}{$\bf 83.98$} \\
\bottomrule
\end{tabular}
\end{center}
\vspace{-5mm}
\caption{Evaluation results on NLVR2.}
\label{tab:nlvr2_result}
\vspace{-3mm}
\end{table*}

In Tables~\ref{tab:vqa_result} to \ref{tab:nlvr2_result}, we report the detailed results for each downstream task, respectively. 
$(\RN{1})$ The \textbf{VQA} results are shown in Table~\ref{tab:vqa_result}, where our single \shortb{} model outperforms the best ensemble model (InterBERT large~\cite{lin2020interbert}) on the VQA leaderboard as of Dec. 12, 2020 \footnote{VQA leaderboard: \url{https://eval.ai/web/challenges/challenge-page/514/leaderboard/1386}}. 
$(\RN{2})$ The \textbf{GQA} results are shown in Table~\ref{tab:gqa_result}, where \short{}w/\vinvl is the first VLP model that outperforms the neural state machine (NSM)~\cite{hudson2019learning} which contains some sophisticated reasoning components deliberately designed for the task.
$(\RN{3})$ The \textbf{Image Captioning} results on the public ``Karpathy" 5k test split are shown in Table~\ref{tab:mscoco_caption}. Table~\ref{tab:coco_leaderboard} shows on a concise version of the COCO image captioning online leaderboard\footnote{Image Captioning Leaderboard: \url{https://competitions.codalab.org/competitions/3221\#results}}. The online testing setting reports the results on 40K images, with 5 reference captions (c5) and 40 reference captions (c40) per image. At the time of submitting this paper, our single model achieves No.1 on the entire leaderboard, outperforming all 263 models, including many ensemble (and anonymous) models.
$(\RN{4})$ The Novel Object Captioning (\textbf{NoCaps}) results are shown in Table~\ref{tab:nocaps_caption}. Without any VLP, i.e. by directly training a BERT-based captioning model on COCO, the model with our new visual features (denoted as VinVL) already surpasses the human performance 
in CIDEr\footnote{NoCaps leaderboard: \url{https://eval.ai/web/challenges/challenge-page/355/leaderboard/1011}}. By adding VIVO~\cite{hu2020vivo} pre-training, our VinVL improves the original VIVO result by 6 CIDEr points and creates a new SoTA.
$(\RN{5})$ Overall, on all these tasks (VQA in Table~\ref{tab:vqa_result}, Image Captioning in Table~\ref{tab:mscoco_caption}, NoCaps in Table~\ref{tab:nocaps_caption}, Image-Text Retrieval in Table~\ref{tab:mscoco_retrieval}, NLVR2 in Table~\ref{tab:nlvr2_result}), we show that \shortb{} can match or outperform previous SoTA large models, and \shortl{} substantially uplifts the SoTA. 

\subsection{Ablation Analysis}
\label{subsec:vision_ablation}
We select the VQA task for the ablation study because its evaluation metric is well-defined and the task has been used as a testbed for all VLP models. To assist our analysis, we create a local validation set, vqa-dev, out of the standard validation set to select the best model during training for evaluation. 
vqa-dev contains randomly sampled 2K images and their corresponding questions, amounting to 10.4K image-QA pairs in total. 
Except for Table~\ref{tab:overall_result} and \ref{tab:vqa_result}, all our VQA results are reported on this vqa-dev set. 
Unless otherwise specified, the reported STD is half of the difference of two runs of the VQA training with different random seeds. 

In VQA, the VL model $y = \mathbf{VL}(\wv, \qv, \vv)$ has $\wv$ as the question and $y$ as the answer. We focus on studying the effect of visual features $\vv$ produced by different Vision models $\mathbf{Vision}(Img)$ to better understand their relative contribution in the VQA performance. To eliminate the impact of using different tags $\qv$, we use the same tags in the VQA models of \textsc{Oscar}~\cite{li2020oscar}. All the ablation experiments are conducted using models of the BERT-base size.

\paragraph{How much do the V and VL matter to the SoTA?}
\begin{table*}[t!]
\begin{center}
\begin{tabular}{ c|ccc }
 \backslashbox{vision}{vl} & no VLP &
 \shortstack{$\textsc{Oscar}_\text{B}$ \\ \small{\cite{li2020oscar}}}  & \shortstack{\shortb{} \\ \small{(ours)}} \\
\hline
R101-C4~\cite{anderson2018bottom} & 68.52 \small{$\pm$0.11} & 72.38 & 72.46\small{$\pm$0.05}  \\
VinVL (ours) & 71.34 \small{$\pm$0.17} & -- & 74.90\small{$\pm$0.05} \\
\end{tabular}
\end{center}
\vspace{-3mm}
\caption{Effects of vision (V) and vision-language (VL) pre-training on VQA.}
\label{tab:pretrain_ablation}
\end{table*}
\label{subsec:oscar_ablation}
Table~\ref{tab:pretrain_ablation} shows the VQA results with different vision models, i.e., R101-C4 model from \cite{anderson2018bottom} and 
our X152-C4 model pre-trained with 4 datasets (VinVL), and with different VLP methods, i.e., no VLP, \oscar~\cite{li2020oscar} and our \short. 
Taking the $\textsc{Oscar}_\text{B}$ model with R101-C4 features as the baseline, the \shortb{} model with our X152-C4 features improves the absolute accuracy from 72.38 to 74.90, in which the \short{} pre-training contributes 5\% of the gain (i.e., $72.38\rightarrow 72.46$) and the vision pre-training (improved visual features) 95\% (i.e., $72.46\rightarrow 74.90$). This demonstrates that vision representations matter significantly in VLP and downstream tasks.

Taking the ``no VLP'' model with R101-C4 features as the baseline, Table~\ref{tab:pretrain_ablation} shows that the gains of VinVL ($71.34-68.52 = 2.82$) and VLP ($72.46-68.52 = 3.94$) are 
additive ($74.90-68.52 \approx 2.82 + 3.94$). This is intuitive because vision pre-training and VLP improve the Vision model 
$\mathbf{Vision}(Img)$ 
and VL model 
$\mathbf{VL}(\wv, \qv, \vv)$ 
separately. This also indicates that our pre-trained vision model can be utilized in any VL models by directly replacing their vision models, such as R101-C4 \cite{anderson2018bottom}, with ours. 

\paragraph{How much do data and model sizes matter to the new vision model?} The improvement of VQA from R101-C4~\cite{anderson2018bottom} to VinVL (ours) in Table~\ref{tab:pretrain_ablation} is a compound effect of increasing model size (from R101-C4 to X152-C4) and data size (from VG to our merged four OD datasets). Table~\ref{tab:model_data_size} shows the ablation of the two factors without VLP. Although VG's large object and attribute vocabulary allows to learn rich semantic concepts, VG does \emph{not} contain large amounts of annotations for effective training of deep models. 
Vision models trained using the merged four OD datasets perform much better than VG-only-trained models, and the improvement is larger with the increase of the model size.\footnote{The R101-C4 model in Table~\ref{tab:model_data_size} is exactly the VG-pre-pretrained model from \cite{anderson2018bottom}. We do not train this model on our merged OD dataset because this model architecture is old-fashioned and is slow to train.}

\paragraph{How much does OD model architecture matter?} 
The choice of model architecture affects the VQA performance. 
Table~\ref{tab:model_data_size} shows that R50-FPN under-performs R50-C5 when they are trained only on VG; but the performance gap diminishes when both are trained on the merged dataset (4Sets). 
A detailed comparison between FPN and C4 architectures is presented in Appendix~\ref{app:c4vsfpn}.

\begin{table}[t!]
\begin{center}
\begin{tabular}{ c|cccc  }
\backslashbox{data}{model} & R50-FPN & R50-C4 & R101-C4~\cite{anderson2018bottom} & X152-C4 \\
\hline
VG & 67.35\small{$\pm$0.26} & 67.86\small{$\pm$0.31} & 68.52 \small{$\pm$0.11} & 69.10\small{$\pm$0.06} \\
4Sets$\rightarrow$VG & 68.3\small{$\pm$0.11} & 68.39\small{$\pm$0.16} & -- & 71.34 \small{$\pm$0.17} \\
\end{tabular}
\end{center}
\vspace{-4mm}
\caption{Ablation of model size and data size on training vision models.}
\label{tab:model_data_size}
\vspace{-1mm}
\end{table}

\begin{table*}[t]
\begin{center}
\begin{threeparttable}
\begin{tabular}{ c|cc|cc|cc}
Model & \multicolumn{2}{c|}{R50-FPN} & \multicolumn{2}{c|}{R50-C4} & \multicolumn{2}{c}{X152-C4} \\
Pre-training dataset & ImageNet & 4Sets & ImageNet & 4Sets & ImageNet5k & 4Sets \\
\midrule
COCO $mAP$ & 40.2~\cite{wu2019detectron2} & 44.78\tnote{*} & 38.4~\cite{wu2019detectron2} & 42.4 & 42.17 & 50.51 \\
\hline
\shortstack{VG obj $mAP^{50}$ \\ attr $mAP$ with gt boxes}
& \shortstack{9.6\\5.4} & \shortstack{11.3\\5.5} & \shortstack{9.6\\6.3} & \shortstack{12.1 \\ 6.1} & \shortstack{11.2 \\ 6.6} & \shortstack{13.8 \\ 7.1} \\
\bottomrule
\end{tabular}
\begin{tablenotes}
\item[*] Since our four pre-training datasets contain Objects365, it is not surprising that we obtain better results than 42.3 $mAP^{50}$ in \cite{shao2019objects365}, which is obtained by pre-training on Objects365.
\end{tablenotes}
\end{threeparttable}
\end{center}
\vspace{-4mm}
\caption{Effect of vision pre-training on object detection tasks.}
\label{tab:odtasks}
\end{table*}

\paragraph{How much does OD pre-training matter for object detection tasks?} 
Table~\ref{tab:odtasks} presents the object detection results on COCO and the object-attribute detection results on VG (1594 object classes, 524 attribute classes). 
The results show that OD pre-training benefits the object detection tasks. 
Note that the mAP on VG is much lower than that on typical OD datasets (such as COCO) due to two reasons: (1) VG contains a large number of object classes with limited and extremely unbalanced annotations, (2) there are many missing annotations in the VG evaluation data.\footnote{As a reference, the R101-C4 model from \cite{anderson2018bottom} on VG with 1600 objects and 400 attributes has mAP of 8.7/7.8 evaluated in our code, whereas it was reported as 10.2/7.8 due to differences in OD evaluation pipeline.} 
Although the mAP numbers are low, the detection result using X152-C4 is reasonably good; see Appendix~\ref{app:visionqa} for more visualizations.
We also see that FPN models perform consistently worse in attribute detection than C4 models, neither do FPN models show any advantage in object detection on VG. This contributes to the inferior performance of FPN, compared to C4, on downstream VL tasks, as discussed in Section~\ref{subsec:vision_pretrain}.

\paragraph{How much does the diversity of visual concepts, i.e., object and attribute vocabularies, matter?}
\begin{table*}[t!]
\begin{center}
\begin{tabular}{ c|c|cccc|c}
Dataset name & 
ImageNet & 
VG-obj & 
VG w/o attr & 
VG~\cite{anderson2018bottom} & 
VG & 
4Sets$\rightarrow$VG \\
\#obj \& \#attr & 
1000 \& 0 & 
317 \& 0 & 
1594 \& 0 & 
1600 \& 400 & 
1594 \& 524 & 
1848 \& 524 \\
\midrule
R50-C4 + BERT$_{B}$ &
66.13\small{$\pm$0.04} &
64.25\small{$\pm$0.16} &
66.51\small{$\pm$0.11} &
67.63\small{$\pm$0.25} &
67.86\small{$\pm$0.31} &
68.39\small{$\pm$0.16} \\
\end{tabular}
\end{center}
\vspace{-4mm}
\caption{Effect of object-attribute vocabulary. We use all grid features (maximal 273) for the ImageNet classification model (first column), and maximal 50 region features for OD models (other columns).}
\label{tab:od_objattr}
\end{table*}
We directly train vision models 
on different datasets, including 
(1) standard ImageNet with 1K classes (ImageNet), 
(2) Visual Genome with 317 object classes (VG-obj) that are shared with COCO 80 classes and OpenImagesV5 500 classes,
(3) VG with all 1594 object classes (VG w/o attr), 
(4) VG with 1594 object classes and 524 attribute classes (VG), and 
(5) the merged OD dataset (4Sets) for pre-training and VG for fine-tuning.
For all the OD models (the last four columns in Table~\ref{tab:od_objattr}), we initialize the OD training with an ImageNet-pre-trained classification model, and use maximal 50 \textit{region} features per image as input to the VL fusion module. 
For the ImageNet pre-trained classification model (the second column in Table~\ref{tab:od_objattr}), we use all the \textit{grid} features (maximal 273) for each image\footnote{Our use of grid feature follows PixelBert~\cite{huang2020pixel}. See Appendix~\ref{appsec:gridfeature} for details.}. 
The results show that
\begin{itemize}[noitemsep,leftmargin=*,topsep=2pt]
    \item In general, vocabularies with richer objects lead to better VQA results: 
    VG-obj $<$ ImageNet $<$ VG w/o attr. The VG-obj vocabulary contains 79 of 80 COCO classes (only missing \texttt{potted plant}) and 313 of 500 OpenImagesV5 classes, and is a good approximation of common object classes of typical OD tasks. However, our results show that this vocabulary is not rich enough for VL tasks because it misses many important visual concepts (e.g., \texttt{sky}, \texttt{water}, \texttt{mountain}, etc.) which are crucial for VL tasks, as also illustrated by the comparison of detected regions in Figure~\ref{fig:pretrain_finetune}.
    \footnote{Using the same training procedure on VG, we trained an R50-C4 model on the OpenImagesV5 dataset (500 classes). Using the region features produced by this model, the VQA performance is 63.55$\pm$0.14. The result is slightly worse than that of VG-obj because both VG and VQA images are from the COCO dataset but OpenImages images are not.
    }.
    \item Attribute information is crucial to VL tasks: models trained with attributes (VG and 4Sets$\rightarrow$VG) are significantly better than those without attributes.
    \item Even for the small vision model R50-C4, vision pre-training improves visual features for VQA, i.e., 4Sets$\rightarrow$VG is the best performer.
\end{itemize}

In Table~\ref{tab:region_effect}, we use different kinds of region proposals to extract image features. COCO groundtruth object regions (GT-Obj, 80 classes) and object-stuff regions (GT-Obj\&Stuff, 171 classes) are perfect in terms of localization, but their vocabulary sizes are limited. Regions proposed by VG-trained models ([2] and VinVL) are imperfect in localization but using a larger vocabulary. For the VQA task, COCO GT boxes are much worse than the proposals generated by VG-trained models. The result demonstrates the difference between the typical OD tasks and the OD tasks in VL: OD in VL requires much richer visual semantics to align with the rich semantics in the language modality. This further echoes our claim that an image understanding module trained using richer vocabularies performs better for VL tasks.
\begin{table}[ht]
\begin{center}
{\begin{tabular}{ c|cccc  }
 \backslashbox{model}{region} & GT-Obj & GT-Obj\&Stuff & \shortstack{Anderson \\ et al. [2]} & VinVL (ours) \\
\hline
\shortstack{Anderson \\ et al. [2]} & 63.81 {$\pm$0.94} & 66.68 {$\pm$0.16} & 68.52 {$\pm$0.11} & 69.05 {$\pm$0.06} \\
VinVL (ours) & 65.60 {$\pm$0.21} & 68.13 {$\pm$0.26} & 70.25 {$\pm$0.05} & 71.34 {$\pm$0.17} \\
\end{tabular}}
\end{center}
\vspace{-4mm}
\caption{Effect of different region proposals on VQA.}
\label{tab:region_effect}
\vspace{-4mm}
\end{table}

\section{Conclusion}
\label{sec:con}
In this paper we have presented a new recipe to pre-train an OD model for VL tasks.
Compared to the most widely used \emph{bottom-up and  top-down} model \cite{anderson2018bottom}, the new model is bigger, better-designed for VL tasks, and pre-trained on much larger text-image corpora, and thus can generate visual features 
for a richer collection of visual objects and concepts that are crucial for VL tasks.
We validate the new model via a comprehensive empirical study where we feed the visual features to a VL fusion model which is pre-trained on a large-scale paired text-image corpus and then fine-tuned on seven VL tasks. 
Our results show that the new OD model can substantially uplift the SoTA results on all seven VL tasks across multiple public benchmarks.
Our ablation study shows that the improvement is mainly attributed to our design choices regarding diversity of object categories, visual attribute training, training data scale, model size, and model architecture.



\section*{Acknowledgement}
We thank Xi Yin for her contributions to this project while she was in Microsoft. We thank Xiyang Dai for his conjecture that C4 arch is better than FPN because C4 arch makes better use of ImageNet initialization weights. 

{\small
\bibliographystyle{ieee_fullname}
\bibliography{egbib}
}

\appendix
\clearpage
\newpage
\section{Qualitative study of three pre-trained vision models}
\label{app:visionqa}
\begin{figure*}[t!]
\includegraphics[width=0.99\textwidth]{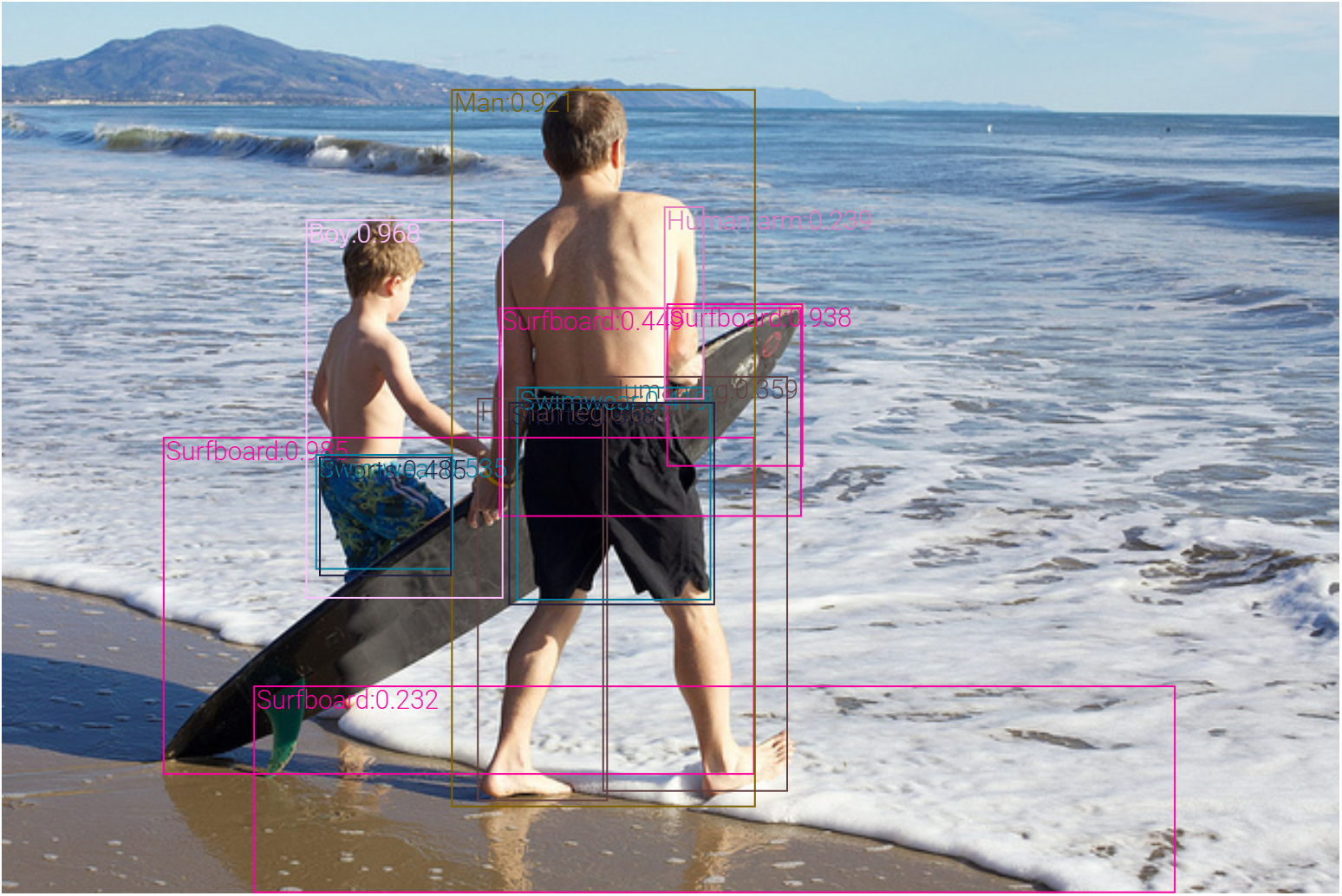}
\caption{Predictions from X152-FPN trained on OpenImages. Test image: COCO\_test2015\_000000028839}
\label{fig:oi_pretrain_finetune}
\end{figure*}
\begin{figure*}[t!]
\centering
\includegraphics[width=0.89\textwidth]{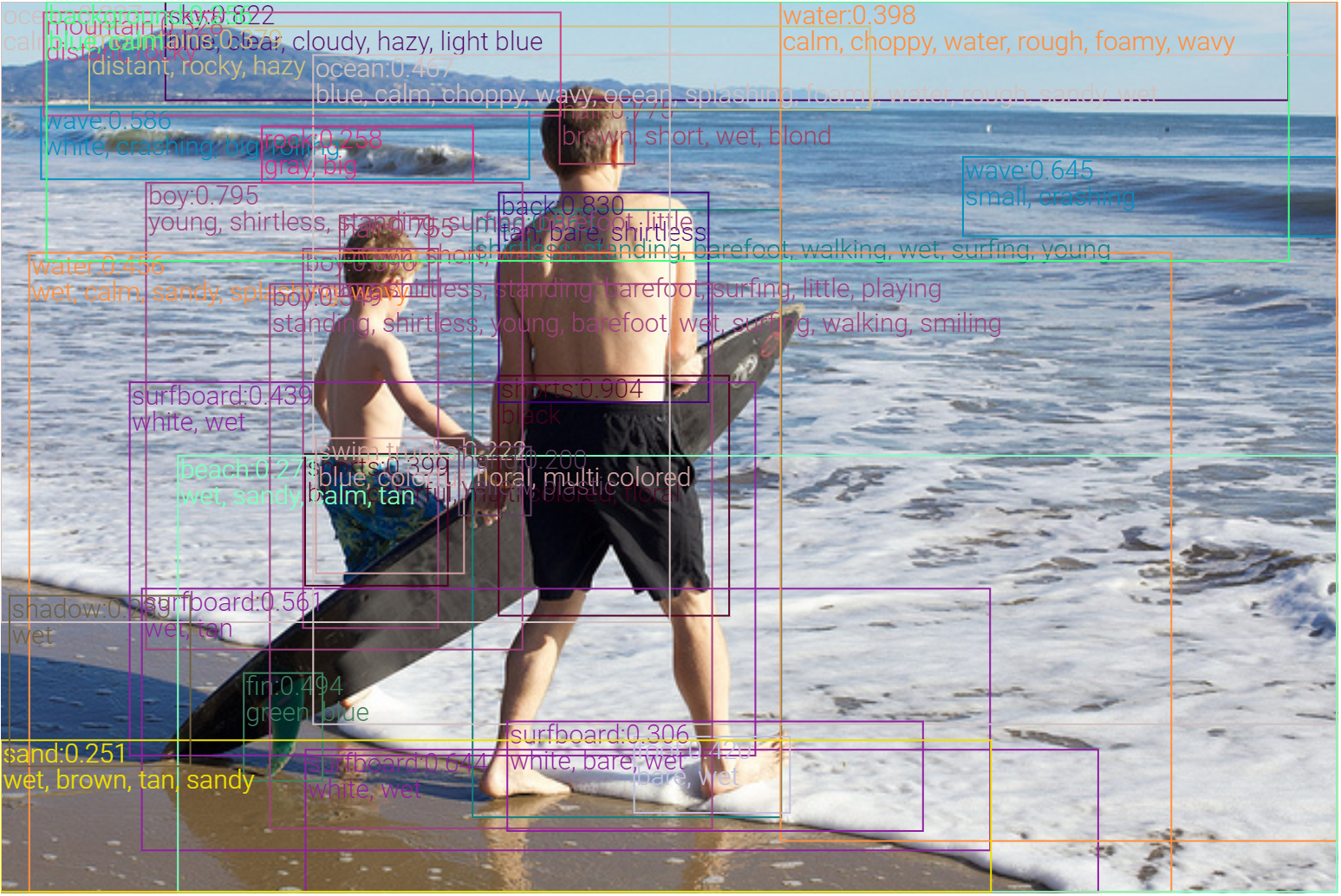}\\
\includegraphics[width=0.89\textwidth]{vinvl_figs/VinVL_X152C4.PNG}
\caption{Predictions from R101-C4 trained on VG from~\cite{anderson2018bottom} (top), X152-C4 pre-trained on 4 OD datasets and finetuned on VG (bottom). Test image: COCO\_test2015\_000000028839}
\label{fig:vg_pretrain_finetune}
\end{figure*}

We apply three (pre-trained) object detection models on the image in Figure \ref{fig:pretrain_finetune} and list their detection results for a more detailed comparison.

Detections from X152-FPN trained on Open Images V5. See Figure \ref{fig:oi_pretrain_finetune}:\\
\texttt{Surfboard; Surfboard; Surfboard; Surfboard; Man; Human leg; Human leg; \\ Swimwear; Swimwear; Shorts; Shorts; Boy; Human arm}.

Detections from R101-C4 trained on VG by Anderson \etal ~\cite{anderson2018bottom}. There are obviously wrong detections, marked in \textcolor{red}{red}. See Figure \ref{fig:vg_pretrain_finetune} (top):\\
\texttt{black shorts; young, shirtless, standing, barefoot, surfing, little, \\playing boy; shirtless, standing, barefoot, walking, wet, surfing, young man; tan, bare, shirtless back; blue, clear, cloudy, hazy, light blue sky; young, shirtless, standing, surfing, barefoot, little boy; brown, short, wet, blond hair; brown, short, wet, blond hair; small, crashing wave; \\ white, wet surfboard; white, crashing, big, rolling wave; \\ wet, tan surfboard; green, blue fin; blue, calm, choppy, wavy, ocean, \\splashing, foamy, water, rough, sandy, wet ocean; wet, calm, sandy, \\splashing, wavy water; white, wet surfboard; bare, wet foot; \\blue, colorful, multi colored, floral shorts; calm, choppy, water, rough, foamy, wavy water; distant, rocky, hazy mountains; standing, shirtless, \\young, barefoot, wet, surfing, walking, smiling boy; calm ocean; distant, rocky mountain; white, bare, wet surfboard; wet, sandy, calm, tan beach; \textcolor{red}{gray, big rock}; \textcolor{red}{blue, calm background}; wet, brown, tan, sandy sand; \\wet shadow; blue, colorful, floral, multi colored swim trunks; \\ \textcolor{red}{yellow, plastic hand}}.

Detections from our pre-trained X152-C4 model pre-trained on four datasets and fine-tuned on VG. There are some repetitive detections, but no obvious wrong detections. See Figure \ref{fig:vg_pretrain_finetune} (bottom): \\
\texttt{blue, green fin; young, barefoot, shirtless, standing, surfing, smiling, little, playing, looking, blond boy; young, barefoot, standing, shirtless, smiling, surfing, blond, playing, looking, little, walking, riding boy; \\ shirtless, barefoot, standing, young, smiling, surfing, walking, wet, playing man; bare, wet foot; black, white surfboard; small, large, white, crashing, big, water, rolling, splashing, rough, foamy wave; bare, wet foot; dark, black, wet, cast shadow; blue, clear, hazy, cloudy, cloudless sky; black, gray, white, raised surfboard; black, wet, short short; brown, short, blond, wet, curly, wavy hair; distant, brown, large, rocky, hazy, big mountain; brown, short, dark, blond, wet hair; blue, white, calm, wavy, choppy, ocean, splashing, water, rough, clear, shallow water; bare, tan, light, beige back; black, blue, wet surfboard; small, dark, water, crashing, rolling, splashing, big wave; wet, white, sandy, tan surfboard; blue, colorful, floral, multi colored, patterned trunk; wet, brown, sandy, tan sand; white, blue, calm, foamy, choppy, splashing, wavy, ocean, rough, water, clear, shallow water; wet, brown, sandy, calm, tan, shallow, smooth, muddy, rough beach; black, white, young board; shirtless, young, standing, barefoot, smiling, surfing, looking, walking, playing boy; blue, calm, choppy, wavy, ocean, clear, rough, splashing, water, foamy, shallow, rippled ocean; yellow, gold bracelet; white, silver, black logo; wet, bare, bent, tan, crossed, hairy, short, skinny, back, muscular, extended, outstretched leg; black, gray, white board; brown, distant, large, rocky, big hill; brown, short, blond, wet, curly head; red, black logo; bare, raised, extended, holding, open, up, bent, outstretched hand; black, wet swim trunks; bare, wet, bent, tan, crossed, skinny, short, back, muscular leg; wet, brown, muddy, sandy, tan, shallow reflection}.

\section{\short{} pre-training}
\subsection{Pre-training Corpus}
\begin{table*}[ht!]
\begin{center}
\resizebox{\linewidth}{!}{
\begin{tabular}{c|c|c|c|c|c|c|c|c}
\toprule
Small & \multicolumn{5}{c|}{0.22M Images, 2.5M QAs, 0.7M captions}  \\
\midrule
Medium & \multicolumn{6}{c|}{ 1.89M Images, 2.5M QAs, 0.7M captions, 1.67M pseudo-captions} \\
\midrule
Large & \multicolumn{8}{c}{ 5.65M Images, 2.5M QAs, 4.68M captions, 1.67M pseudo-captions} \\
\midrule
\multirow{2}{*}{Source} & VQA & GQA & VG-QA & COCO & Flicker30k & OpenImages & CC & SBU \\
 & (train) & (bal-train) & (train) & (train) & (train) & (od train) & (train) & (all) \\ 
\midrule
Image/Text & 83k/545k & 79k/1026k & 87k/931k & 112k/559k & 29k/145k & 1.67M/1.67M & 3.1M/3.1M & 875k/875k \\
\midrule
$\wv,\qv,\vv$ & \multicolumn{3}{c|}{ Question, Answer, ImageFeatures} & \multicolumn{5}{c}{(Generated) Caption, (Generated) ImageTags, ImageFeatures} \\
\bottomrule
\end{tabular}}
\end{center}
\caption{Statistics of the pre-training corpus.}
\label{tab:pretrain_corpus}
\end{table*}
Table~\ref{tab:pretrain_corpus} shows the statistics of image and text of the pre-training corpora. In our ablation study, we use corpora of three different sizes: `Small', `Medium', `Large'. Different from \oscar~\cite{li2020oscar}, we make use of image tagging datasets OpenImages, by generating captions using \oscar's image captioning model to form triplets of (generated caption, image tags, image features) for \short{} pre-training. By self-training technique, our pre-training corpora can be scaled to a much larger amount by making use of large-scale image tagging datasets, e.g., OpenImages (9M) and YFCC (92M).

\subsection{\short{} pre-training objectives}
\label{app:oscarlosses}
\paragraph{Masked Token Loss: A Loss Mimics Image Captioning.} 
The word tokens of image captions (questions) $\wv$ and word tokens of object tags (answers) $\qv$ share the same linguistic semantic space, and the Masked Token Loss (MTL) is applied on tokens of both $\wv$ and $\qv$. 
We define the {\it discrete token sequence} as $\hv \triangleq [\wv, \qv]$, and apply the Masked Token Loss (MTL) for pre-training. At each iteration, we randomly mask each input token in $\hv$ with probability $15\%$, and replace the masked one $h_i$ with a special token $\mathtt{[MASK]}$. The goal of training is to predict these masked tokens based on their surrounding tokens $\hv_{\backslash i}$ and image features $\vv$ by minimizing the negative log-likelihood:
\begin{align}
\hspace{-0mm}
\Lcal_{\text{MTL}} = -\E_{  (\vv, \hv) \sim \Dcal } \log p( h_i | \hv_{\backslash i}, \vv )
\label{eq_attend_mlm}
\end{align}
This is the same MTL as in \oscar~\cite{li2020oscar} and similar to the masked language model used by BERT. The masked word or tag needs to be recovered from its surrounding context, with additional image information to help ground the learned word embeddings in the vision context.

\begin{figure*}[t!]
\centering
\includegraphics[width=0.8\textwidth]{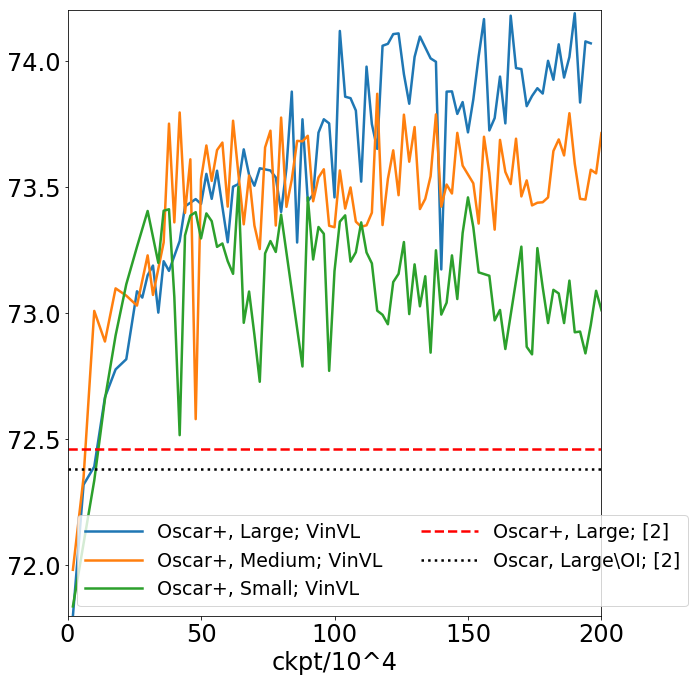}
\caption{Effect of \short{} pre-training corpus size and effect of self-training by making use of tagging data in \short{}. Each curve, with legend ``VLP, Corpus; VisionFeature'', denotes a VLP experiment where the VLP method is either \oscar \, or \short{}, the VLP pre-training Corpus is Small/Medium/Large (defined in Table~\ref{tab:pretrain_corpus}), and VisionFeature is either our new vision features (VinVL for short) or those from \cite{anderson2018bottom} (\cite{anderson2018bottom} for short). X-axis denotes the pre-training iterations of \short{} checkpoints. Y-axix is the vqa-dev accuracy of a VQA model initialized from the corresponding pre-training checkpoint and fine-tuned with a fixed scheme. Compared with ``\short{}, Small; VinVL" (green), ``\short{}, Medium; VinVL" (yellow) adds the 1.7M OpenImages Tagging data into the pre-training and its performance gets improved significantly, demonstrating the effect of self-training by making use of tagging data. The ``\short{}, Large; VinVL" (blue) further scales up the pre-training corpus by adding Google Conceptual Captions and SBU datasets with generated tags and its performance gets further improved, demonstrating the effect of \short{} pre-training corpus size. As baselines, we also provide performance of \oscar\, and \short{} with image features from \cite{anderson2018bottom}, which clearly demonstrates that our new image features (VinVL) matter significantly in the VL pre-training and VL downstream tasks.}
\label{fig:selftraining_ablation}
\end{figure*}

\paragraph{3-way Contrastive Loss: A Loss Mimics Text-Image Retrieval and Visual Question Answering Simultaneously.} We present our 3-way contrastive loss in Section~\ref{subsec:oscarobjective} in the main paper.

\subsection{Ablation of the two new techniques}
\paragraph{Effect of self-training: Leveraging Image Tagging data.} In Figure~\ref{fig:selftraining_ablation}, we show the effect of self-training by making use of tagging data in \short{}, by fine-tuning \short{} pre-training checkpoints on VQA. Compared with ``\short{}, Small; VinVL" (green), ``\short{}, Medium; VinVL" (yellow) adds the 1.7M OpenImages Tagging data into pre-training and its performance gets improved significantly, demonstrating the effect of self-training by making use of tagging data. As baselines, we also provide performance of \oscar\, and \short{} with image features from \cite{anderson2018bottom}, which clearly demonstrates that the new image features pre-trained by VinVL matter significantly in the VL pre-training and VL downstream tasks. 

\paragraph{Effect of the new 3-way contrastive loss.} As illustrated in Table~\ref{tab:contrastive}, with the new 3-way contrastive loss, the VQA performance is the same as the \oscar \ pre-training, while the Text-Image Retrieval performance improves significantly compared with the \oscar \ pre-training.

\paragraph{Overall improvement from \oscar ~to \short{}.} We point out that the improvement from \oscar ~to \short{} with image features from \cite{anderson2018bottom} is minor, because (1) we only add 1.7M OpenImages' tagging data
to enlarge the pre-training corpus, which is a small portion compared with \oscar's original pre-training corpus (i.e., Large$\backslash$OI, 3.98M images and 7.18M image-caption pairs), and (2) the new 3-way contrastive loss has more significant improvements in Text-Image Retrieval tasks than that in the VQA task, as illustrated in Table~\ref{tab:contrastive}. We would expect much more significant improvements when we scale up the \short{}'s pre-training corpus to a much larger scale by adding large scale image tagging datasets, e.g., OpenImages (9M) and YFCC (92M).


\section{Downstream Tasks Fine-tuning}
\label{sec:downstreams}
We follow the downstream task fine-tuning recipes in \oscar~\cite{li2020oscar}.
\subsection{VQA}
Given an image and a question, the task is to select the correct answer from a multi-choice list, it requires the model to answer natural language questions based on an image. Here we conduct experiments on the widely-used VQA v2.0 dataset~\cite{goyal2017making}, which is built on the MSCOCO~\cite{lin2014microsoft} images. 
Following~\cite{anderson2018bottom}, for each question, the model picks the corresponding answer from a shared set of $3,129$ candidates.

When fine-tuning on the VQA task, the input sequence contains the concatenation of a given question, object tags and object region features, and then the $\mathtt{[CLS]}$ output from \short{} is fed to a task-specific linear classifier for answer prediction.
Similarly as the literature~\cite{anderson2018bottom}, we treat VQA as a multi-label classification problem – assigning a soft target score to each answer based on its relevancy to the human answer responses, and then we fine-tune the model by minimizing the cross-entropy loss computed using the predicted scores and the soft target scores. During inference, we simply use Softmax for answer prediction.

For VQA training, we random sample a set of 2k images from the MS COCO validation set as our validation set, the rest of images in the training and validation are used in the VQA fine-tuning. For the \shortb{} model, we fine-tune for $25$ epochs with a learning rate of $5e^{-5}$ and a batch size of $128$. For the \shortl{} model, we fine-tune for $25$ epochs with a learning rate of $3e^{-5}$ and a batch size of $96$.

\subsection{GQA}
Similarly as VQA, GQA tests the reasoning capability of the model to answer a question. We conduct experiments on the public GQA dataset~\cite{hudson2019gqa}. 
For each question, the model chooses an answer from a shared set of $1,852$ candidates. Our fine-tuning procedure is following Oscar~\cite{li2020oscar,chen2019meta}, which first fine-tunes the model on unbalanced ``all-split'' for $5$ epochs with a learning rate of $5e^{-5}$ and a batch size of $128$, and then fine-tuned on the ``balanced-split'' for $2$ epochs.


\subsection{Image Captioning}
An image captioning model generates a natural language description for a given image. To enable sentence generation, we fine-tune \short{} using the seq2seq objective. The input samples are processed to triples consisting of image region features, captions, and object tags, in the same way as that during the pre-training. We randomly mask out $15\%$ of the caption tokens and use the corresponding output representations to perform classification to predict the token ids. Similar to previous works~\cite{li2020oscar,zhou2019unified}, the self-attention mask is constrained such that a caption token can only attend to the tokens before its position to simulate a uni-directional generation process.  
Note that all caption tokens will have full attentions to image regions and object tags but not the other way around. 

During inference, we first encode the image regions, object tags, and a special token $\mathtt{[CLS]}$ as input. Then the model starts the generation by feeding in a $\mathtt{[MASK]}$ token and 
selecting
a token from the vocabulary based on the likelihood output. Next, the $\mathtt{[MASK]}$ token in the previous input sequence is replaced with the selected token and a new $\mathtt{[MASK]}$ is appended for the next word prediction. The generation process terminates when the model outputs the $\mathtt{[SEP]}$ token. We use beam search (\ie beam size = 5)~\cite{anderson2018bottom} in our experiments and report our results on the COCO image captioning dataset. 

Though the training objective (\ie seq2seq) for image captioning is different from that used in pre-training (\ie bidirectional attention-based masked token loss), we directly fine-tune \short{} for image captioning on COCO without additional pre-training on Conceptual Captions~\cite{sharma2018conceptual}. This is to validate the generalization ability of the \short{} models for generation tasks. We use the same Karpathy split~\cite{karpathy2015deep}. 
For the \shortb{} model, we fine-tune with cross-entropy loss for $30$ epochs with a batch size of $256$ and an initial learning rate of $1e^{-5}$ and then with CIDEr optimization~\cite{rennie2017self} for $10$ epochs with a batch size of $128$ and initial learning rate of $2e^{-6}$.
We compare with several existing methods, including
BUTD~\cite{anderson2018bottom},
VLP~\cite{zhou2019unified},
AoANet~\cite{huang2019attention},
OSCAR~\cite{li2020oscar}.

\subsection{NoCaps}
Novel Object Captioning~\cite{agrawal2019nocaps} extends the image captioning task, is to test models' capability of describing novel objects from the Open Images dataset~\cite{kuznetsova2018open} which are not seen in the training corpus. Following the restriction guideline of NoCaps, 
we train \short{} on COCO without the initialization from pre-training, so no additional image-text pairs are used for training except COCO. 

Since NoCaps images are collected from Open Images, we train an object detector using the Open Images training set and apply it to generate the tags. We conduct experiments from BERT model directly without pre-training as required by the task guidelines. For the \shortb{} model, we train $30$ epochs with a batch size of $256$ and learning rate $1e^{-4}$; further we perform CIDEr optimization with learning rate $5e^{-6}$ and batch size $112$ for $10$ epochs. During inference, we use constrained beam search for decoding. We compare \short{} with 
OSCAR~\cite{li2020oscar} on this task.

\subsection{Image-Text Retrieval}
There are two sub-tasks: {\it image retrieval} and {\it text retrieval}, depending on which modality is used as the retrieved target. Both tasks calculate a similarity score between an image and a sentence, which heavily relies on the cross-modal representations.
%

Following Oscar~\cite{li2020oscar}, we formulate the retrieval as a binary classification problem, where given an aligned image-text pair, we randomly select a different image or a different sentence to form an unaligned pair. The final representation of $\mathtt{[CLS]}$ is used as the input to the classifier to predict whether the given pair is aligned or not. 
In the testing stage, the probability score is used to rank the given image-text pairs of a query.

Following~\cite{li2019unicoder}, we report the top-$K$ retrieval results on both the $1$K and $5$K COCO test sets.
We adopt the widely used Karpathy split~\cite{karpathy2015deep} on the COCO caption dataset~\cite{lin2014microsoft} to conduct our experiments. Specifically, the dataset consists of $113,287$ images for training, $5,000$ images for validation, and $5,000$ images for testing. Each image is associated with $5$ human-generated captions. For the \shortb{} model, we fine-tune with a batch size of $256$ for $40$ epochs. The initial learning rate is set to $2e^{-5}$ and linearly decreases. For the \shortl{} model, we fine-tune with a batch size of $128$ for $40$ epochs. The initial learning rate is set to $1e^{-5}$ and linearly decreases. We use the validation set for parameter tuning. 
We compare with several existing methods, including
DVSA~\cite{karpathy2015deep},
VSE++~\cite{faghri2017vse++}, 
DPC~\cite{zheng2017dual}, 
CAMP~\cite{wang2019camp}, 
SCAN~\cite{lee2018stacked}, 
SCG~\cite{shi2019knowledge}, 
PFAN~\cite{wang2019position}, 
Unicoder-VL~\cite{li2019unicoder},
12-in-1~\cite{lu201912},
UNITER~\cite{chen2019uniter}.

\subsection{NLVR2}
Given a pair of images and a natural language, the goal of NLVR2~\cite{suhr2018corpus} is to determine whether the natural language statement is true about the image pair. 
%
For NLVR2 fine-tuning, we first construct two input sequences, each containing the concatenation of the given sentence (the natural language description) and one image, and then two $\mathtt{[CLS]}$ outputs from \short{} are concatenated as the joint input for a binary classifier, implemented by an MLP.

For the \shortb{} model, we fine-tune for $20$ epochs with learning rate \{$2e^{-5}$, $3e^{-5}$, $5e^{-5}$\} and a batch size of $72$. For the \shortl{} model, we fine-tune for $20$ epochs with learning rate of \{$2e^{-5}$, $3e^{-5}$\} and a batch size of $48$.


\begin{figure}[t!]
\centering
\includegraphics[width=0.7\columnwidth]{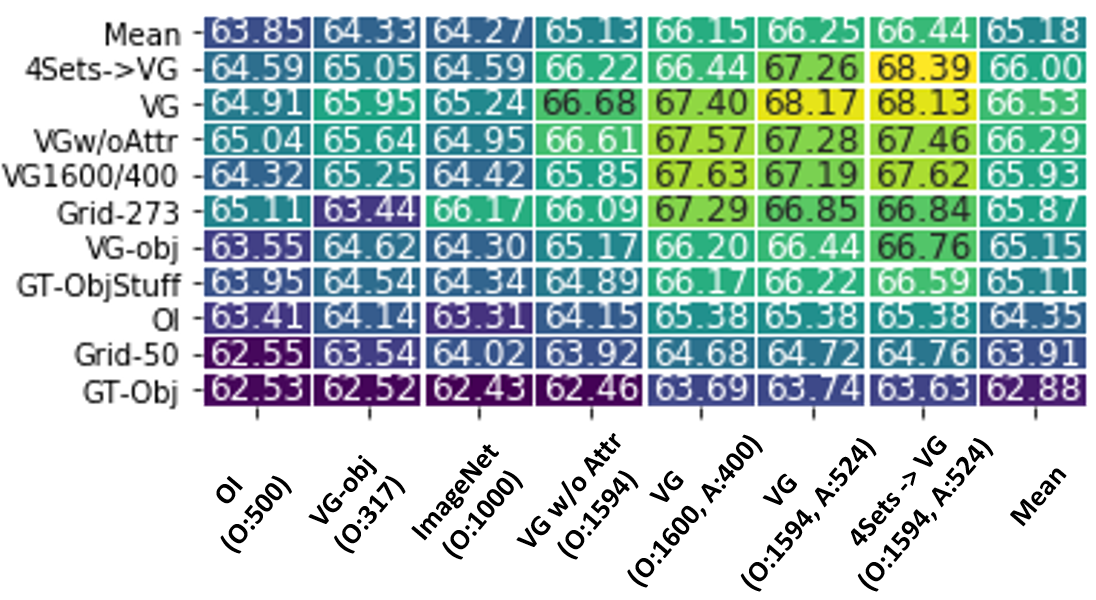}
\caption{Overall comparison of vocabulary effect on VQA. X-axis: how the R50-C4 model is trained; Y-axis: how the feature is extracted (grid or region features, different kinds of boxes to extract region features). All region features have maximal 50 regions. The top row ``Mean'' is the average over all rows, showing the overall quality of different vision models. The far-right column ``Mean'' is the average over all columns, showing the overall quality of different feature extraction methods.}
\label{fig:vocab_overall}
\end{figure}

\begin{figure*}[t!]
\centering
\includegraphics[width=0.49\textwidth]{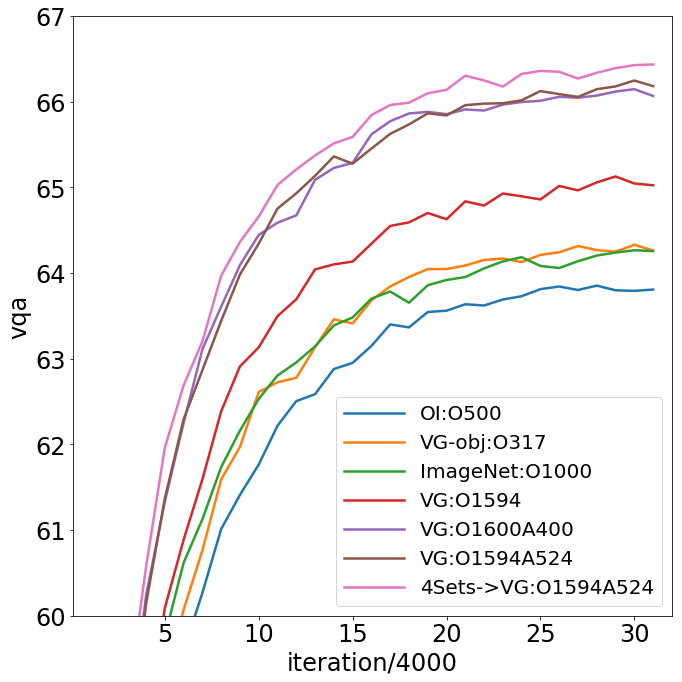}
\includegraphics[width=0.49\textwidth]{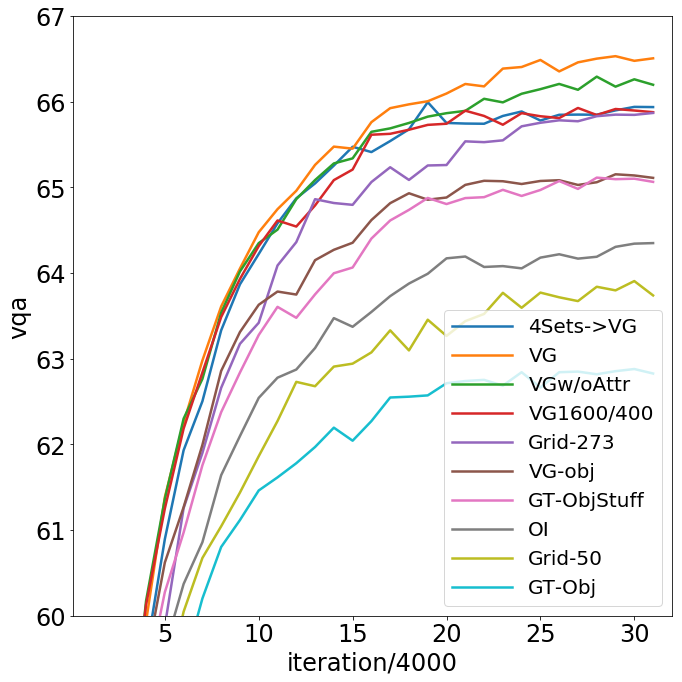}
\caption{Left: comparison of object vocab and attribute vocab, average over all types of bounding boxes. Right: comparison of feature extraction methods, average over all types of pre-trained vision models. X-axis is the number of iterations when we take the checkpoint for evaluation. Y-axis is the VQA accuracy on our vqa-dev.}
\label{fig:vocab_boxtypes}
\end{figure*}

\section{More on the Effect of the Object-Attribute Vocabulary Size: disentangling the effects of region proposals and model weights}
In Section~\ref{subsec:vision_ablation}, we demonstrate that the more diverse the visual concepts (object and attribute vocabularies) are, the better the visual region features for VL tasks. The better performance may come from the more diverse proposed regions where the region features are extracted (see the comparison in Figure~\ref{fig:pretrain_finetune}, ``region" for short), or from the better model weights that can produce better high-dimensional region representation even for the same region (``model" for short). In this section, we disentangle effects of region proposals and model weights, by performing synthetic experiments in which we use region proposals from one vision model and model weights from another vision model. Our results show that both the region proposals and model weights matter for VL tasks.

\subsection{Disentangling the effects of region proposals and model weights on R50-C4}
As in Section~\ref{subsec:vision_ablation}, We train vision models $\vv = \mathbf{Vision}(Img)$ on different datasets, i.e., OpenImages with 500 object classes (OI:O500), standard ImageNet with 1K classes (ImageNet:O1000), Visual Genome with 317 object classes (VG-obj), Visual Genome with 1594 object classes (VG:O1594), VG with 1594 object classes and 524 attribute classes (VG:O1594A524), pretrain on the merged 4 datasets and finetune on VG:O1594A524 (4Sets$\rightarrow$VG:O1594A524). For {\it each} model, we also try different ways to extract features: (1) \textit{region} features from different models' proposed regions (same notations with models) where each image has maximal 50 region features, and (2) \textit{grid} features where we use all grid features (Grid-273) or randomly sampled 50 grid features (Grid-50) for each image. We present the results of these model-region cross-combination experiments in Figure~\ref{fig:vocab_overall}. We also present the mean accuracy over all box types to obtain a robust ranking of different checkpoints and the mean accuracy over all checkpoints to obtain a robust ranking of different box types. We have the following observations:
\begin{itemize}[noitemsep,leftmargin=*,topsep=2pt]
    \item The richer the object vocabulary is, the better for VQA: OI:500 $\approx$ VG-obj:O317 $<$ ImageNet:O1000 $<$ VG:O1594. 
    \item Attribute information is crucial to VL tasks: all features trained with attributes (Columns with VG:O1594A524) are significantly better than those without attributes.
    \item Even for small vision backbone R50, vision pre-training makes vision features better: Column \\ ``4Sets$\rightarrow$VG:O1594A524" are better than all other columns. Notice that the vision pre-training improves both the region features and the grid features.
    \item It is crucial to extract features from semantically diverse regions: regions from OI and VG-obj are significantly worse than all other regions, and is even worse than grid features.
    \item Grid features perform worse than region features with regions proposed by VG models. By comparing Row ``Grid-273'' with rows with VG regions, it seems hopeful to close this gap while paying more hardware memory and computational cost in cross-modal models $\mathbf{VL}$. It is three times slower to {\bf train} the ``Grid-273'' models than training models with region features. 
\end{itemize}

In Figure~\ref{fig:vocab_boxtypes}, instead of just showing one final number, we provide the {\it mean evaluation curves along training trajectories} to demonstrate the ranking, as an even more robust evidence. These results further confirm the conclusions we draw in Section~\ref{subsec:vision_ablation}.  

\subsection{Disentangling the effects of region proposals and model weights on the SoTA model}
In Table~\ref{tab:regionmodel_ablation}, we alternate the combination of region proposals and model weights, and evaluate them on VQA. As we can see, the improvement of using boxes from the R101-C4 model~\cite{anderson2018bottom} to extract features from our X152-C4 model is much bigger than that of using boxes from our X152-C4 model to extract features from the R101-C4 model~\cite{anderson2018bottom}, indicating pre-trained model weights are more important than regions. Inspired by this analysis, we propose the class-agnostic NMS for region selection in the box head of the OD model, which does not sacrifice any VQA performance but greatly improves the model's inference speed. This analysis also suggests that large-scale OD pre-training should improve performance for grid-feature based VL models, as supported by more results in Appendix~\ref{appsec:gridfeature}. 

In Table~\ref{tab:regionmodel_ablation}, We also report VQA results with COCO groundtruth object regions (GT-Obj, 80 classes) and object-stuff regions (GT-Obj\&Stuff, 171 classes).
For VQA task, COCO GT boxes are much worse than proposals from VG trained models. This shows the difference between typical OD tasks and OD in VL: OD in VL requires much richer visual semantics to align with the rich semantics in the language modality. This further echoes with our claim that an image understanding module trained with rich semantics is crucial for VL tasks.

\begin{table}[ht]
\begin{center}
{\begin{tabular}{ c|cccc  }
 \backslashbox{model}{region} & GT-Obj & GT-Obj\&Stuff & \shortstack{Anderson \\ et al. [2]} & VinVL (ours) \\
\hline
\shortstack{Anderson \\ et al. [2]} & 63.81 {$\pm$0.94} & 66.68 {$\pm$0.16} & 68.52 {$\pm$0.11} & 69.05 {$\pm$0.06} \\
VinVL (ours) & 65.60 {$\pm$0.21} & 68.13 {$\pm$0.26} & 70.25 {$\pm$0.05} & 71.34 {$\pm$0.17} \\
\end{tabular}}
\end{center}
\vspace{-4mm}
\caption{Ablation of region and model on VQA.}
\label{tab:regionmodel_ablation}
\vspace{-4mm}
\end{table}

\section{More on FPN and Comparison of C4 and FPN}
\label{app:c4vsfpn}
\subsection{Two reasons why FPN performs worse than C4 on VL tasks.}
\begin{table*}[t!]
\begin{center}
\resizebox{\linewidth}{!}{
\begin{tabular}{ c|cccccc  }
 & \shortstack{no image feature \\ {\small $\wv$}} & \shortstack{R50-C4 {\small w/ box head} \\ {\small randomly initialized}} & R50-FPN & R50-C4 & 4Sets$\rightarrow$ R50-FPN & 4Sets$\rightarrow$R50-C4 \\
\hline
VG-trained & -- & 67.6 \small{$\pm 0.13$} & 67.6\small{$\pm 0.30$} & 68.0\small{$\pm 0.16$} & 68.3\small{$\pm 0.11$} & 68.2\small{$\pm 0.05$} \\
Initial & 55.5\small{$\pm 0.50$} & 61.8 \small{$\pm 0.47$} & 57.6\small{$\pm 0.16$} & 64.8\small{$\pm 0.44$} & 66.1\small{$\pm 0.23$} & 66.8\small{$\pm 0.21$} \\
\end{tabular}
}
\end{center}
\caption{C4 vs FPN architecture on VQA. Boxes used to extract features $\vv$ and tags $\qv$ used in $\mathbf{VL}$ model are the same with those used in \textsc{Oscar}~\cite{li2020oscar}. Row ``Initial'' means using the initialization model without VG training for feature extraction.}
\label{tab:c4vsfpn}
\end{table*}
Our experimental results confirm the conclusion of \cite{jiang2020defense} that the FPN model does not provide better region features for VL tasks than the C4 model (Columns ``R50C4'' vs. ``R50FPN'' in Table~\ref{tab:c4vsfpn}). 
Our analysis reveals two reasons. 
First of all, all layers involved in feature extraction in the C4 model have been pre-trained using ImageNet while the MLP head of FPN does not. It turns out that the VG dataset is still small to train a good visual features for VL tasks and using ImageNet-pre-trained weights is beneficial. This can be verified by  two experiments: (1) When the R50-C4 model is trained on VG with its box head randomly initialized (VG-trained - R50C4 w/ box head randomly initialized), the C4 model's performance is the same as FPN;  and (2) C4 and FPN achieve the same performance after vision pre-training on 4 datasets (68.3 vs. 68.2).  
The second reason is due the network architecture (CNN vs. MLP) of the box head in the OD model. The convolutional head in C4 has a better inductive bias in encoding visual information than the MLP head in FPN. This can be verified by the fact that when vision features from randomly initialized models are used (Row ``Initial'' in Table~\ref{tab:c4vsfpn}), R50-C4 performs much better than R50-FPN, indicating that the initial C4 features encode much more useful visual information than the inital FPN features. 
The ``random'' C4 features nearly match the feature from ImageNet pre-trained model (Row ``Initial'' Column ``R50C4''), while ``random'' FPN features are close to the performance without visual features as input (Row ``Initial'' Column ``no image feature $\wv$'').

\subsection{Effect of pooling methods in FPN on VQA performance.}
\begin{figure*}[t!]
\centering
\includegraphics[width=0.49\textwidth]{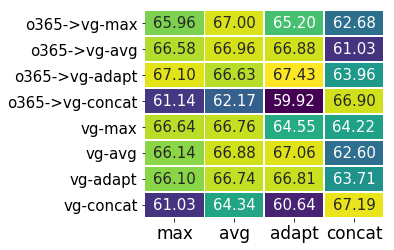}
\includegraphics[width=0.49\textwidth]{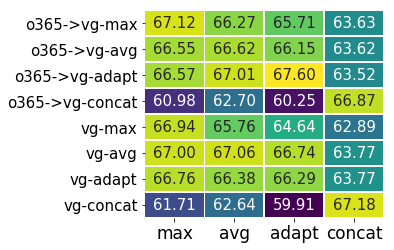}
\caption{Pooling methods in FPN feature extraction are not the root cause of FPN's worse performance than C4. X-axis: the pooling method when extracting features for VL tasks; Y-axis: the pooling method (vision model) when pre-training the visual feature extraction model. All experiments are using regions from the Bottum-up Top-down model~\cite{anderson2018bottom}. Each combination is experimented twice with two random seeds, i.e. seed=42 on the left and seed=88 on the right. The results from two random seeds are consistent.}
\label{fig:fpn_feature_pooltype_seed42}
\end{figure*}

Different from C4 models that extract region features from a single scale (the end of C4 block), FPN models extract region features from multiple scales adaptively based on the area of the region. Therefore, there is some in-homogeneity in FPN's region features since they may come from different scales. In Figure~\ref{fig:fpn_feature_pooltype_seed42}, we show that this is not the cause of FPN's worse performance than C4 on the VQA task. More specifically, we experiment with 4 pooling methods for FPN architecture. (1) adapt: the original FPN's pooling method that extract features adaptively from different scales; (2) max: extract features from all scales and then do a max-pool; (3) avg: extract features from all scales and then do an average-pool; (4) concat: extract features from all scales and then concatenate them together. We also train multiple FPN models on VG with these pooling methods, with or without pre-training on the Objects365 dataset. We experiment on all possible combinations (in total $8 \times 4$) of 8 vision models and 4 pooling methods on the VQA task. When there is a parameter dimension mis-match, e.g., non-concat FPN models but use concat pooling methods in VQA and vice versa, we specify those parameter randomly with PyTorch's default initialization method. The results in Figure~\ref{fig:fpn_feature_pooltype_seed42} shows that (1) there is no obvious difference in different pooling methods, with the default ``adapt'' and the ``concat'' methods perform slightly better than ``max'' and ``avg''; (2) (without surprise) the performance is significantly worse when there is a parameter dimension mismatch between vision models and VL task feature extraction methods, i.e., non-concat FPN models but use concat pooling methods in VQA and vice versa. These results show that the pooling method (no matter in vision model training or in VL task feature extraction) is not the root cause of FPN's worse performance than C4 on the VQA task.

\begin{figure*}[t!]
\centering
\includegraphics[width=0.49\textwidth]{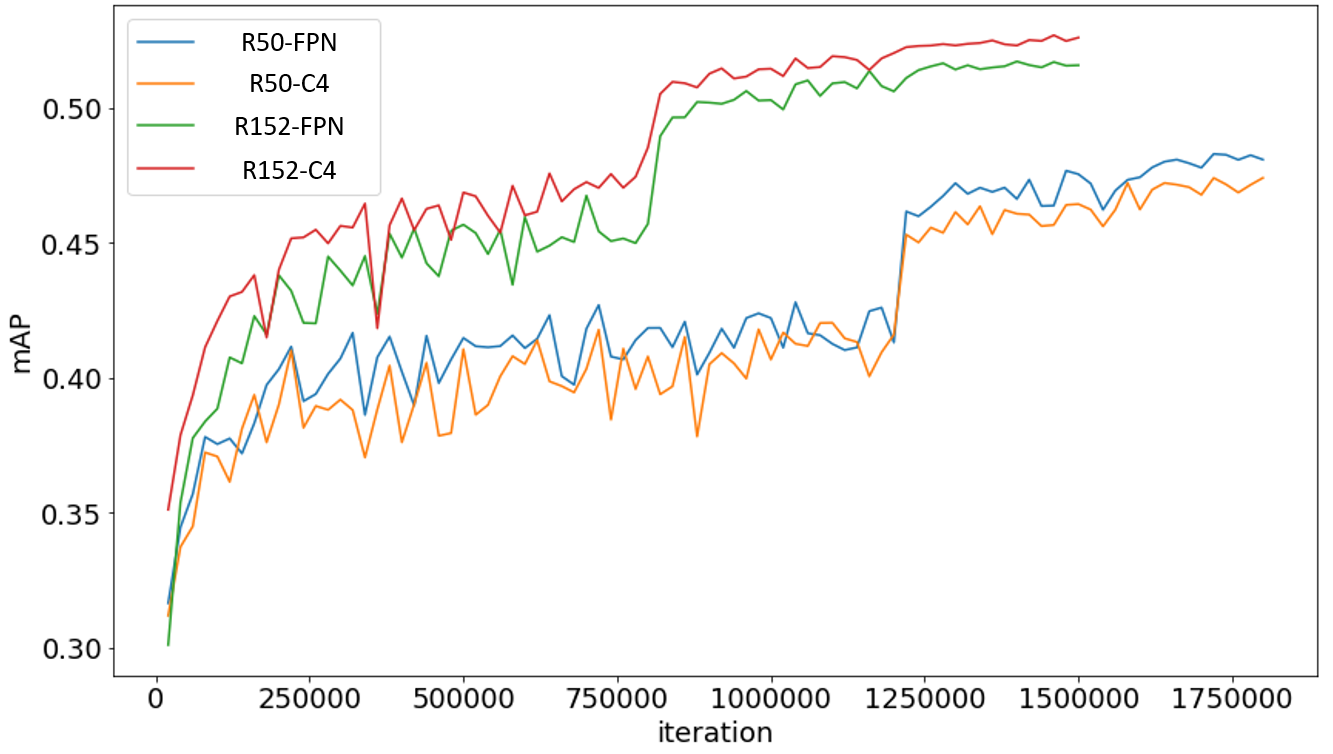}
\includegraphics[width=0.49\textwidth]{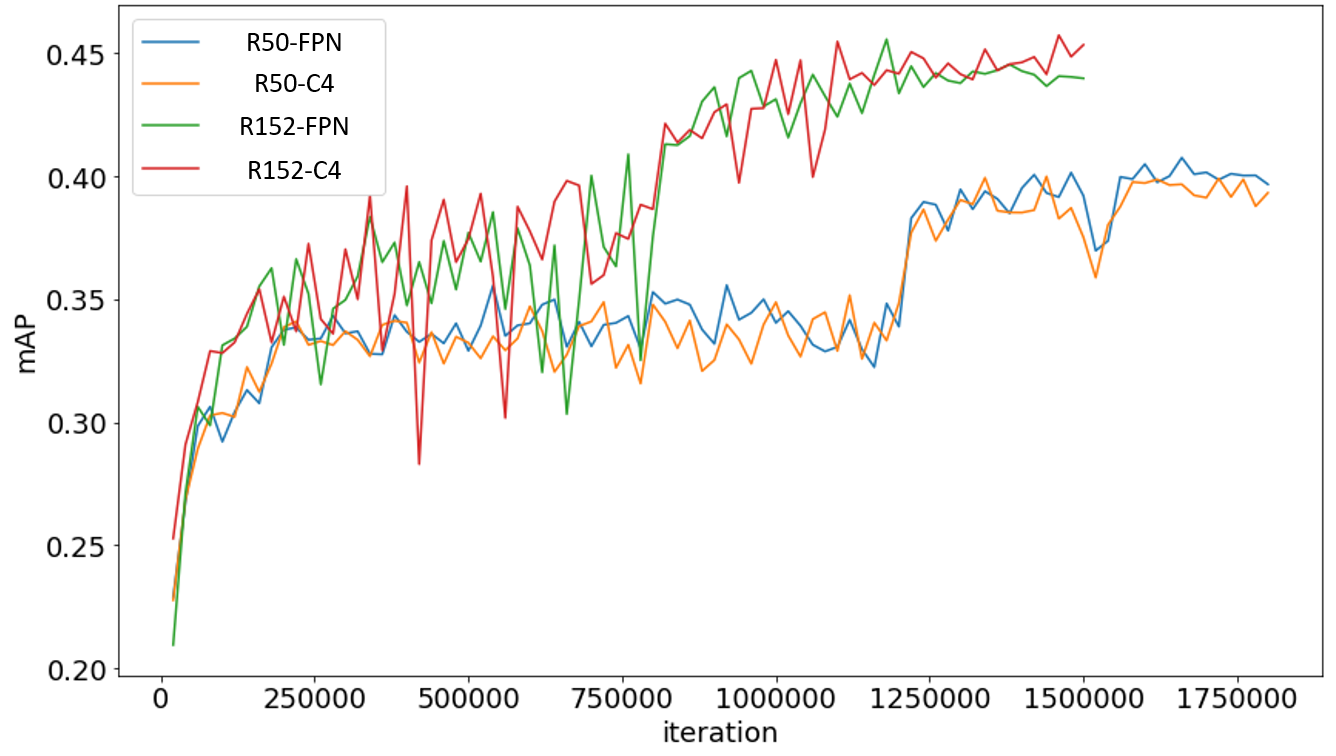} \\
\includegraphics[width=0.49\textwidth]{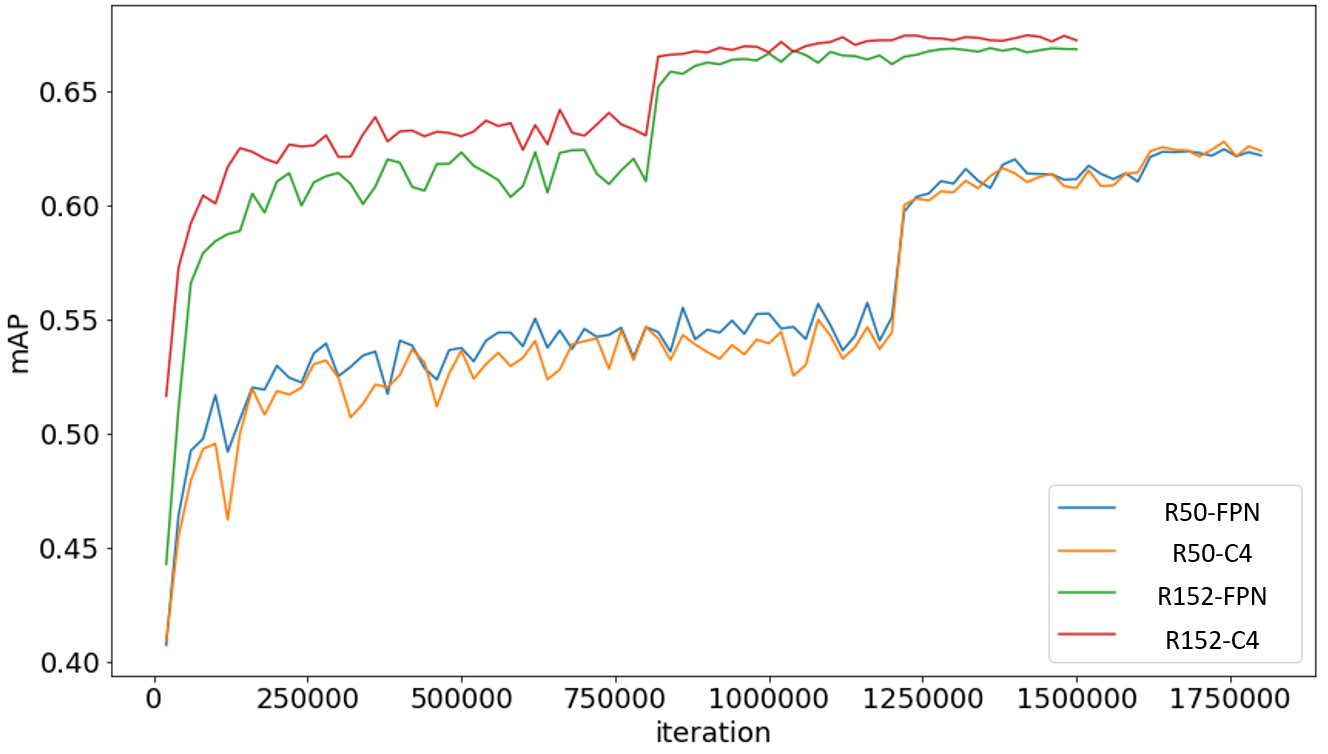}
\includegraphics[width=0.49\textwidth]{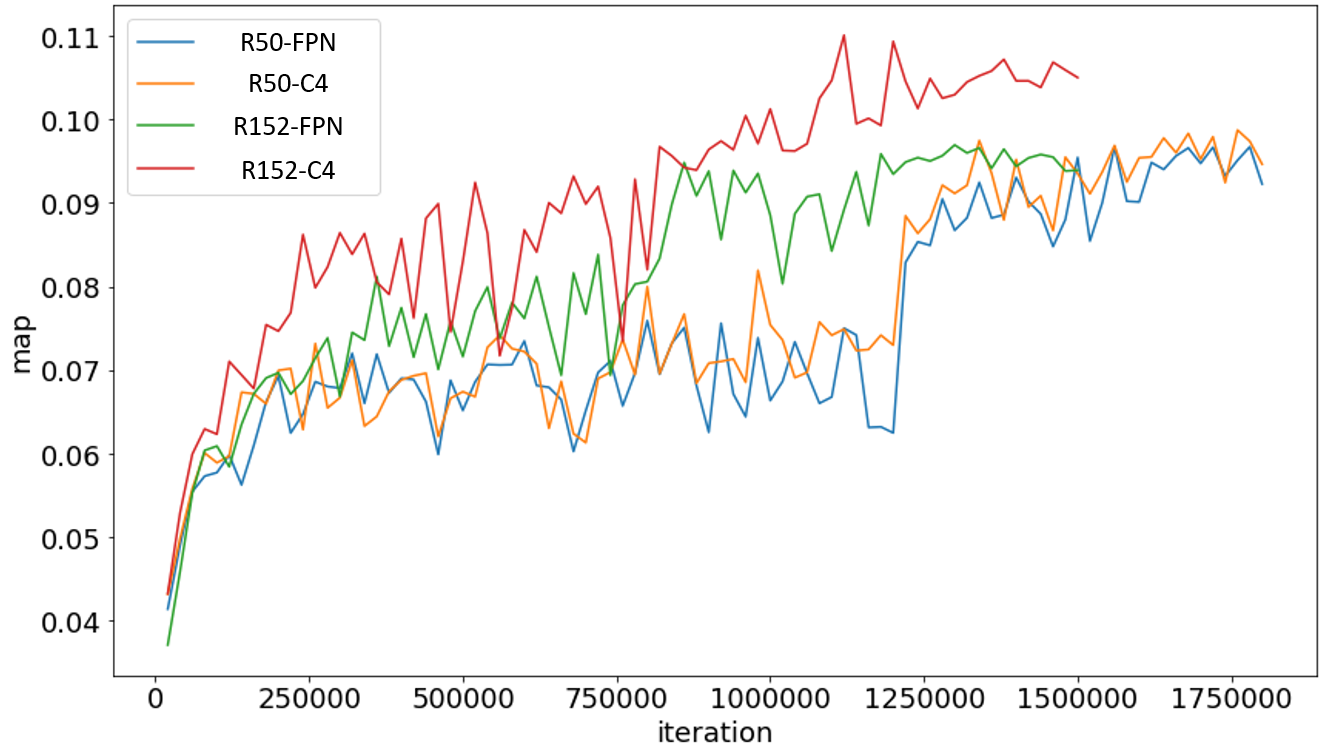}
\caption{Checkpoints' $mAP^{50}$ on 4 validation sets: COCO with stuff (top left), Objects365 (top right), OpenImages (bottom left) and Visual Genome (1594 object classes, bottom right). For R50 models, the R50-FPN model is slightly better than C4 on COCO and Objects365 but slightly worse than C4 on Visual Genome. For R152 models, the R152-FPN model is consistently worse than the R152-C4 model on all 4 different datasets.}
\label{fig:4setsmap_fpnvsc4}
\end{figure*}

\begin{table*}[t!]
\begin{center}
\begin{threeparttable}
\begin{tabular}{ c|cccc  }
\toprule
 & \shortstack{ImageNet-5k \\ {\cite{wu2019detectron2}}} & 4Sets & VG with Attr & 4Sets$\rightarrow$VG \\
\hline
grid feature (273) & 68.3\small{$\pm 0.29$} & 65.2\small{$\pm 2.47$} & 67.5\small{$\pm 0.20$} & 69.4\tnote{*} \\
region feature (50) & 67.7\small{$\pm 0.16$} & 68.5\small{$\pm 0.13$} & 69.8\small{$\pm 0.23$} & 70.6\small{$\pm 0.13$} \\
\bottomrule
\end{tabular}
\begin{tablenotes}
\item[*] The other run failed and thus there is no std for this experiment.
\end{tablenotes}
\end{threeparttable}
\end{center}
\caption{Ablation study of X152 models on VQA. Vision models in the last three columns are trained with initialization from the ImageNet-5k checkpoint in the first column. All the region features are extracted with boxes proposed by our best X152-C4 model (pre-trained on 4Sets and fine-tuned on VG). By comparing the first column and the last column, we see that our proposed vision pre-training (first on 4 sets and then on VG with attributes) improves performance for both the grid-feature based model and the region-feature based model. Since the X152 backbone is much larger than the R50 backbone in Figure~\ref{fig:vocab_overall}, the larger model can make better use of the large pre-training datasets and thus have more significant improvements.}
\label{tab:X152C4_ablation}
\end{table*}

\subsection{Large-scale object-detection pre-training of C4 and FPN models}
In this paper, we have trained R50-C4, R50-FPN, R152-C4 and R152-FPN models on the merged object detection datasets described in Table~\ref{tab:vision_pretrain_corpus}. In Figure~\ref{fig:4setsmap_fpnvsc4}, we report the $mAP^{50}$ of checkpoints from these 4 experiments on 4 validation sets: COCO with stuff (top left), Objects365 (top right), OpenImages (bottom left) and Visual Genome (1594 object classes, bottom right). For R50 models, the R50-FPN model is slightly better than C4 on COCO and Objects365 but slightly worse than C4 on Visual Genome. For R152 models, the R152-FPN model is consistently worse than the R152-C4 model on all 4 different datasets. Therefore, we finally use the R152-C4 model for downstream vision-language tasks. 

\section{Grid feature}
\label{appsec:gridfeature}
In Table~\ref{tab:X152C4_ablation}, we train grid-feature based and region-feature based X152 models for VQA, with the vision models pre-trained on different vision datasets, i.e., ``ImageNet-5k'' from \cite{wu2019detectron2}, our 4-dataset merged OD dataset~\ref{tab:vision_pretrain_corpus} (4Sets), our VG dataset with 1594 object classes and 524 attribute classes (VG with Attr), and first 4Sets and then VG (4Sets$\rightarrow$VG). Vision models in the last three cases are trained with initialization from the same ImageNet-5k checkpoint from \cite{wu2019detectron2}. All the region features are extracted with boxes proposed by our best X152-C4 model (pre-trained on 4Sets and fine-tuned on VG). By comparing ``ImageNet-5k'' and ``4Sets$\rightarrow$VG'', we see that our proposed vision pre-training improves performance for both the grid-feature based model and the region-feature based model. Since the X152 backbone is much larger than the R50 backbone in Figure~\ref{fig:vocab_overall}, the larger model makes better use of the large pre-training datasets and thus has more significant improvements. It is interesting to see that for grid-feature based models, the ``ImageNet-5k'' model performs better than the ``4Sets'' model and the ``VG with Attr'', while it is not the case for region-feature based models. This may indicate that how the vision model is trained (grid-feature wise or region-feature wise) may have big impact on the downstream VL tasks. 

\begin{table*}[ht!]
\begin{center}
\begin{tabular}{ l|cc|cc|cc }
\multirow{2}{*}{Model} & \multicolumn{2}{c|}{R50-C4} & \multicolumn{2}{c|}{R101-C4~\cite{anderson2018bottom}} & \multicolumn{2}{c}{X152-C4} \\
& $\mathbf{Vision}$ & $\mathbf{VL}$ & $\mathbf{Vision}$ & $\mathbf{VL}$ & $\mathbf{Vision}$ & $\mathbf{VL}$ \\
\hline
Grid-50 & 0.059$\pm$\tiny{0.018} & 0.029$\pm$\tiny{0.002} & 0.083$\pm$\tiny{0.025} & 0.030$\pm$\tiny{0.003} & 0.355$\pm$\tiny{0.022} & 0.031$\pm$\tiny{0.003}  \\
Grid-273 & 0.056$\pm$\tiny{0.005} & 0.027$\pm$\tiny{0.002} & 0.082$\pm$\tiny{0.022} & 0.034$\pm$\tiny{0.001} & 0.344$\pm$\tiny{0.036} & 0.037$\pm$\tiny{0.004} \\
Object & 0.373$\pm$\tiny{0.040} & 0.031$\pm$\tiny{0.005} & 0.663$\pm$\tiny{0.042} & 0.034$\pm$\tiny{0.003} & 0.687$\pm$\tiny{0.064} & 0.036$\pm$\tiny{0.005} \\
Object-eff & 0.165$\pm$\tiny{0.029} & 0.029$\pm$\tiny{0.002} & 0.442$\pm$\tiny{0.119} & 0.036$\pm$\tiny{0.003} & 0.475$\pm$\tiny{0.049} & 0.037$\pm$\tiny{0.005} \\
\midrule
Grid-50 (cpu) & 1.943$\pm$\tiny{0.244} & 0.480$\pm$\tiny{0.042} & 4.050$\pm$\tiny{0.398} & 0.469$\pm$\tiny{0.046} & 17.765$\pm$\tiny{1.693} & 0.501$\pm$\tiny{0.047}  \\
Grid-273 (cpu) & 2.032$\pm$\tiny{0.230} & 1.368$\pm$\tiny{0.056} & 4.052$\pm$\tiny{0.372} & 1.283$\pm$\tiny{0.067} & 17.664$\pm$\tiny{1.713} & 1.326$\pm$\tiny{0.053} \\
Object (cpu) & 11.808$\pm$\tiny{1.322} & 0.500$\pm$\tiny{0.045} & 31.863$\pm$\tiny{7.932} & 0.585$\pm$\tiny{0.044} & 29.641$\pm$\tiny{3.097} & 0.565$\pm$\tiny{0.044} \\
Object-eff (cpu) & 11.729$\pm$\tiny{1.280} & 0.510$\pm$\tiny{0.044} & 31.791$\pm$\tiny{8.027} & 0.587$\pm$\tiny{0.043} & 29.687$\pm$\tiny{3.011} & 0.574$\pm$\tiny{0.036} \\
\end{tabular}
\end{center}
\caption{Time cost of end-to-end inference on VQA. All cross-modal models are BERT-Base. On the SOTA number obtained with X152-C4 region features, the performance {\it keeps the same} when changing to the efficient way to extract the feature while the efficiency greatly improves on GPU. The efficient version does not lead to time saving on CPU, because nearly all inference time is taken by the backbone and C4 head and the time from NMS operations is nearly ignorable on CPU.}
\label{tab:model_efficiency_cpu}
\end{table*}

\section{End-to-end inference efficiency}
\label{app:efficiency}
We report the end-to-end inference time of different VQA models on a Titan-X GPU and a Xeon E5 CPU in Table~\ref{tab:model_efficiency_cpu}. For CPU evaluation, we force that the inference use only one CPU thread. The input image size is $800\times 1333$, and we run the inference with batch size 1 (one image-question pair per batch). We can see that (1) vision models dominate the inference time, especially for large models; (2) models based on grid-feature are faster than those based on region feature; (3) with our proposed fast inference trick, region-feature models are greatly sped up and their inference time can be brought to within 3 times of that of grid-feature models on GPU. We find that on CPU with a single thread, our class-agnostic trick does not lead to time saving, because nearly all inference time is taken by the backbone and C4 head and the time from NMS operations is nearly ignorable on CPU.

\end{document}